\documentclass[acmtog,authorversion,nonacm]{acmart}

\usepackage{booktabs} %

\usepackage[ruled]{algorithm2e} %

\SetAlFnt{\small}
\SetAlCapFnt{\small}
\SetAlCapNameFnt{\small}
\SetAlCapHSkip{0pt}

\usepackage[normalem]{ulem}
\usepackage{stackengine}
\usepackage{bm}
\usepackage[ruled]{algorithm2e} %
\usepackage{wrapfig}
\usepackage[export]{adjustbox}
\usepackage{tabularx}
\usepackage{tikz}
\usepackage{booktabs}
\usepackage{multirow}
\usepackage{tabularx}

\newcommand\hide[1]{}

\newcommand\review[1]{#1}

\definecolor{dbcolor}{RGB}{0,150,200}
\definecolor{lccolor}{RGB}{10,10,210}
\definecolor{pccolor}{RGB}{10,210,10}
\definecolor{gzcolor}{RGB}{150,150,10}
\definecolor{escolor}{RGB}{190,80,90}
\definecolor{bscolor}{RGB}{170,0,230}

\newcommand\ie{\textit{i.e.},~}
\newcommand\eg{\textit{e.g.},~}
\newcommand\Fig[1]{Fig.~\ref{fig:#1}}
\newcommand\Sec[1]{Sec.~\ref{sec:#1}}
\newcommand\Eqn[1]{Eqn.~(\ref{eqn:#1})}

\newcommand\Tab[1]{Table~(\ref{tab:#1})}

\newcommand{\SO}[0]{\mathrm{SO}}

\newcommand{\0}[0]{\mathbf{0}}

\renewcommand{\t}[0]{\mathbf{t}}

\renewcommand{\u}[0]{\mathbf{u}}            %
\renewcommand{\P}[0]{\mathbf{P}}              %
\newcommand{\x}[0]{\mathbf{x}}              %
\newcommand{\X}[0]{\mathbf{X}}      
\newcommand{\A}[0]{\mathbf{A}}              %
\newcommand{\F}[0]{\mathbf{F}}              %
\newcommand{\R}[0]{\mathbf{R}}              %
\newcommand{\D}[0]{\mathbf{D}}              %
\newcommand{\W}[0]{\mathbf{W}}              %
\newcommand{\network}[0]{\mathcal{N}}       %
\newcommand{\simmesh}[0]{\mathcal{M}_{\text{sim}}}
\newcommand{\loss}[0]{\mathcal{L}}
\newcommand{\act}[0]{\mathcal{A}}

\renewcommand{\det}[0]{\operatorname{det}}
\newcommand{\divergence}[0]{\operatorname{div}\cdot \,}
\newcommand{\st}[0]{\operatorname{s.t.} \;}

\newcommand{\skinspace}{\partial \Omega_{\text{skin}}}
\newcommand{\bonespace}{\partial \Omega_{\text{bones}}}
\newcommand{\skullspace}{\partial \Omega_{\text{skull}}}
\newcommand{\jawspace}{\partial \Omega_{\text{jaw}}}

\usepackage{amsmath}

\DeclareMathOperator*{\argmin}{argmin}

\newcolumntype{P}[1]{>{\centering\arraybackslash}p{#1}}

\renewcommand\paragraph[1]{\vspace{2mm}\noindent\textbf{#1}}

\newcommand{\customstrut}[1]{\rule{0pt}{#1}} %

\begin{document}

\title[Learning a Generalized Physical Face Model From Data]
{Learning a Generalized Physical Face Model From Data}

\author{Lingchen Yang}
\orcid{0000-0001-9918-8055}
\affiliation{\institution{ETH Zurich}\country{Switzerland}}
\email{lingchen.yang@inf.ethz.ch}

\author{Gaspard Zoss}
\orcid{0000-0002-0022-8203}
\affiliation{\institution{DisneyResearch|Studios}\country{Switzerland}}
\email{gaspard.zoss@disneyresearch.com}

\author{Prashanth Chandran}
\orcid{0000-0001-6821-5815}
\affiliation{\institution{DisneyResearch|Studios}\country{Switzerland}}
\email{prashanth.chandran@disneyresearch.com}

\author{Markus Gross}
\orcid{0009-0003-9324-779X}
\affiliation{\institution{ETH Zurich}\country{Switzerland}}
\affiliation{\institution{DisneyResearch|Studios}\country{Switzerland}}
\email{gross@disneyresearch.com}

\author{Barbara Solenthaler}
\orcid{0000-0001-7494-8660}
\affiliation{\institution{ETH Zurich}\country{Switzerland}}
\email{solenthaler@inf.ethz.ch}

\author{Eftychios Sifakis}
\orcid{0000-0001-5608-3085}
\affiliation{\institution{University of Wisconsin Madison}\country{USA}}
\email{sifakis@cs.wisc.edu}

\author{Derek Bradley}
\orcid{0000-0002-2055-9325}
\affiliation{\institution{DisneyResearch|Studios}\country{Switzerland}}
\email{derek.bradley@disneyresearch.com}

\renewcommand\shortauthors{L. Yang, G. Zoss, P. Chandran, M. Gross, B. Solenthaler, E. Sifakis, D. Bradley}

\begin{abstract}

Physically-based simulation is a powerful approach for 3D facial animation as the resulting deformations are governed by physical constraints, allowing to easily resolve self-collisions, respond to external forces and perform realistic anatomy edits.  Today's methods are data-driven, where the actuations for finite elements are inferred from captured skin geometry.  Unfortunately, these approaches have not been widely adopted due to the complexity of initializing the material space and learning the deformation model for each character separately, which often requires a skilled artist followed by lengthy network training.  In this work, we aim to make physics-based facial animation more accessible by proposing a generalized physical face model that we learn from a large 3D face dataset.  Once trained, our model can be quickly fit to any unseen identity and produce a ready-to-animate physical face model automatically.  Fitting is as easy as providing a single 3D face scan, or even a single face image.  After fitting, we offer intuitive animation controls, as well as the ability to retarget animations across characters.  All the while, the resulting animations allow for physical effects like collision avoidance, gravity, paralysis, bone reshaping and more.

\end{abstract}


\begin{teaserfigure}
\centering
\includegraphics[width=1\textwidth]{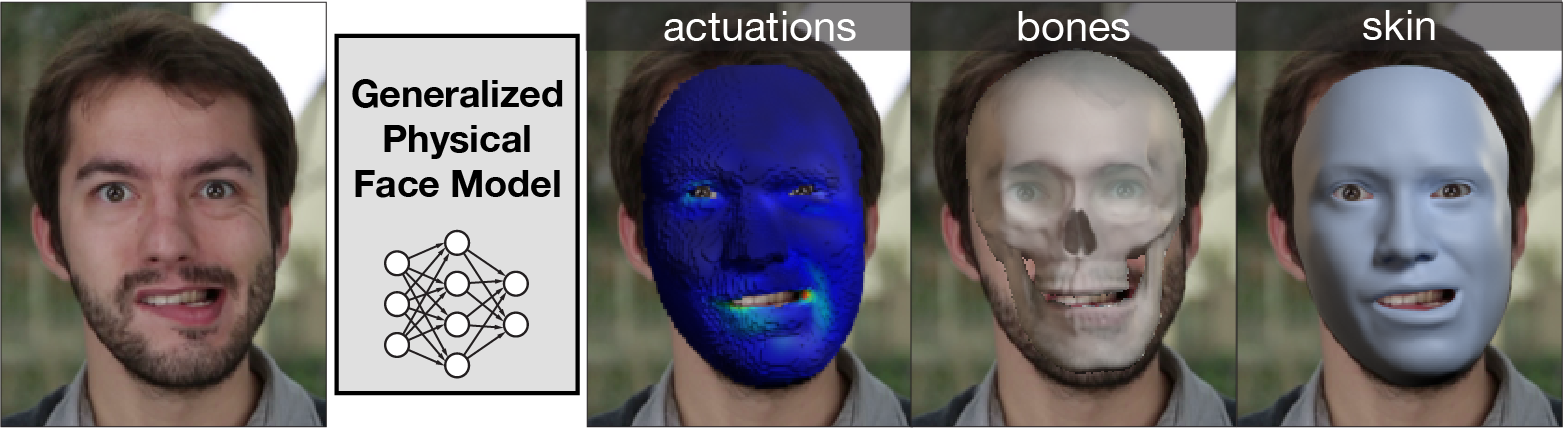}
\caption{
We present a deep generalized physical face model that can be fit to a single face image or 3D scan.  The model produces an identity-specific material space with bones, skin and soft tissue, together with per-expression jaw transformations and element actuations for facial simulation.  Once fit to an unseen identity, the model can be animated to create physics-based facial animation.  Applications like retargeting, interpolation, anatomy editing, and physical effects such as collision avoidance and muscle paralysis are shown in \Sec{Results}.}
\label{fig:teaser}
\end{teaserfigure}

\maketitle

\section{Introduction}
\label{sec:intro}

For decades, 3D facial animation has continued to be a challenging and well-studied problem in computer graphics.  The quest for more realistic facial motion continues because believable 3D characters are one of the leading contributors to widespread adoption of CG content in films, video games and virtual environments.  The reverse is also true, in that unrealistic 3D characters can lead to a widespread rejection of the CG content, which is what makes high-quality facial animation so critical.  Most often, faces are parameterized and controlled by linear blendshape rigs, which are popular due to their simplicity and ease of use.  Unfortunately, blendshape rigs suffer from well-known drawbacks like providing only linear motion and exhibiting surface inter-penetrations (particularly around the lips).  An alternative, and more physically accurate approach is to use physical simulation for facial animation, where anatomical muscle actuations drive the soft tissue deformation.  As a benefit, within a simulator it is easy to add constraints to identify surface contacts and prevent collisions, while producing more natural nonlinear face motion.  Originally, physics-based methods were popular only in academic settings due to their complexity of adoption in production, however recently more anatomical and physics-inspired approaches have gained popularity in the film industry~\cite{Choi2022}\footnote{https://www.fxguide.com/fxfeatured/exclusive-joe-letteri-discusses-weta-fxs-new-facial-pipeline-on-avatar-2/}.

There are generally two approaches to physics-based facial animation. In the first-principles-based method, researchers model the muscle actuation mechanism according to the authentic complex muscle structures in the face, either obtained through MRI~\cite{Sifakis05} or an off-the-shelf muscle template~\cite{ichim2017phace,Bao19}. Although physiologically sound, this strategy is laborious, and its realism highly depends on the accuracy of the anatomical model. As such, a second approach emerged based on shape targeting \cite{klar2020Shapetargeting}, where fine-grained and structure-agnostic actuation is inferred from the captured skin geometry and used to target the undeformed soft tissue~\cite{srinivasan2021learning,yang2022implicit,yang2023ExprAndStyle}. Since only the skin geometry and skeleton are required, this method largely eases the modeling labor while assuming great flexibility, high fidelity, and improved realism.

While the shape targeting approach is attractive, it is currently only tractable for a few hero characters in a production.  This limitation stems from two aspects.  First, although the modeling effort is less than a first-principles approach, there is still a dedicated setup procedure required for each character. 
\review{This involves scanning and tracking extensive facial skin geometry, coupled with capturing the underlying skeleton to delineate the material space—the undeformed soft tissue situated between the rest skeleton and skin, which is essential for simulation. Following this, a dedicated neural network is meticulously trained to predict muscle actuation for each expression-dependent deformation. } 
Secondly, there is a time-consuming, memory-intensive, and identity-specific differentiable simulation \review{based on the Finite Element Method (FEM)} in the training loop.  
As a result, it would be infeasible to learn \review{a universal generic model} from multitudinous identities within a reasonable amount of time and resources, using the currently-available architectures for soft-tissue-based physical face animation. \review{This situation is further aggravated by the fact that the existing large facial datasets lack precise information on the material space, directly forestalling the FEM simulation.}

In this work we aim to alleviate these issues and make physics-based facial animation more freely accessible.  To this end, we propose a new, {\em generalized} physical face model that can be easily adapted to any new character without manual setup costs.  We accomplish this by learning a single implicit neural model for actuation mechanisms trained on a large dataset of hundreds of identities performing a variety of expressions. Naturally, the above-mentioned limitations would pose a considerable problem for this approach as each identity would require a personalized material space and training on such a large dataset with simulation in the loop would be impossible.  We overcome these hurdles with our two main contributions.  First, we propose a design that allows for \review{training without FEM simulation in the loop}, which is fast and memory efficient, allowing us to train on a large dataset of faces and learn the generalizability one would need for such an application.  Second, we propose an architecture that automatically predicts an identity-conditioned material space by warping a single canonical material space. Despite not having \review{FEM} simulation in the training loop nor hand-crafted per-identity material spaces, our network still produces outputs that are compatible with a physics simulator.  The resulting faces can be animated using physics-based facial animation, \review{where various physical effects can be easily incorporated.}

Our model is conditioned on two latent codes, one for identity and one for expression, and the output is an identity-specific material space (\ie skull, jaw, skin and soft tissue in between) coupled with identity- and expression-specific actuations and bone kinematics that can all be readily provided to an off-the-shelf physics simulator.  As a result, our pre-trained network can be used to generate an animation-ready physics-based simulation model for any new character, simply by modifying the identity latent code.  We demonstrate that our model can be fit to a single 3D neutral scan of an actor, or even simply to a single face image.  Once fitted, the model allows animation through the controls of a common 3D morphable face model, which are mapped to our latent space.  Furthermore, our method supports animation retargeting by swapping identity codes.  In all cases, the resulting animation benefits from physical effects like the detection of surface contacts and collision avoidance, the ability to paralyze parts of the face, edit anatomical bone structures, and obey gravitational or other external forces.

In summary, we propose a physical face model that is as generalizable and controllable as current linear blendshape models, but with the added benefit of more physically-accurate facial deformations.  We believe our work will help to democratize physics-based facial animation, making it as simple as fitting an animation-ready simulation model to a single scan or image.

\section{Related Work}

Facial models used in animation range from simple global linear shape models to complex local models that incorporate the underlying facial anatomy through anatomical constraints or physical simulation. We will focus our discussion on anatomical face models and the generation of 3D face models from monocular input.

\subsection{Physics-Based Facial Animation}

Physics-based facial animation typically employs two main approaches. The first, known as the first-principles-based method, requires users to intricately model the muscle actuation mechanism based on the complex structures of facial muscles. These structures are derived from sources such as CT/MRI scans~\cite{Sifakis05} or pre-existing muscle templates~\cite{ichim2017phace,Bao19}. While this approach ensures physiological accuracy, the generation of such models is labor-intensive and the realism of the resulting animation heavily relies on the precision of the anatomical and biomechanical models.

In response to the challenges posed by the first-principles-based method, a second approach has emerged, known as shape targeting \cite{klar2020Shapetargeting}.
In this methodology, every element within the mesh is considered active and subject to actuation for the purpose of inducing forces that drive the motion of the face or body.
To reproduce identity-specific facial expressions, an actuation mechanism is optimized based on face capture data.
In the work of Srinivasan et al.~\shortcite{srinivasan2021learning}, the integration of neural networks with a comprehensive muscle model facilitated the learning of the muscle actuation mechanism. Subsequent advancements in the field transitioned towards an implicit neural representation, rendering the actuation mechanism more compact and independent of resolution~\cite{yang2022implicit}. Building upon this foundation, Yang et al.~\shortcite{yang2023ExprAndStyle} further extended the methodology by training the neural network on captured data from a small number of individuals simultaneously. This multi-identity training approach enables the model to learn diverse expression styles based on distinct actuation patterns and allows applications such as style retargeting.

Notably, these actuation-based approaches streamline the modeling process by only necessitating skin geometry and bone data for model training. Nevertheless, their heavy reliance on identity-specific data, hand-crafted material spaces and architectures that only allow training on one or few identities at a time represent significant limitations, particularly hindering their generalizability and widespread application in production settings. We address this by introducing a generalized physical face model that can be easily adapted to any new character.

The recent work of Wagner et al.~\shortcite{SoftDECA2023} is the closest in spirit to our work, as we share a common motivation although very different solutions. They propose an extension to linear blendshape models that aims to mimic physics-based facial animation within a linear framework.  Physical simulation is only used to generate training data, not at runtime, and thus they are tied to the prescribed simulation effects present in the  dataset.  For example, they cannot handle lip collisions, which is a primary reason to employ simulation for facial animation.  On the other hand, we propose a nonlinear physics-based face model trained on real capture data, which can be fit to any novel identity and provide a true physical model that employs simulation at runtime to allow any desired physical effect, including lip collision avoidance.

\subsection{Anatomically-Constrained Face Models}

Anatomical constraints on the facial surface are often used to plausibly restrict the range of the skin deformations.
The anatomically-constrained local deformation model, introduced in the context of monocular facial performance capture by Wu et al.~\shortcite{Wu2016}, initially established a connection between the skin surface and anatomical bones. This was achieved by modeling the thickness of the soft tissue between a bone point and the skin surface, together with skin sliding coupled with bulging. %
This anatomically-constrained face model was also beneficial in face modeling applications, helping untrained users to quickly create believable digital characters \cite{Gruber2020}.
In the domain of facial performance retargeting, Chandran et al.~\shortcite{Chandran2022Karacast} employed the same model to confine a retargeted shape within the realm of anatomically plausible shapes specific to a target actor.
An implicit variant of the anatomical face model was presented by Chandran and Zoss~\shortcite{chandran2023anatomically}, which facilitates the learning of a continuous anatomical structure that densely constrains the skin surface. The model can disentangle deformation arising from rigid bone motion and non-rigid deformations created by muscle activations.
Qiu et al.~\shortcite{qiu2022sculptor} learned an anatomical facial shape model from medical imaging data, and presented a morphable model that is able to generate faces that jointly model the skull, facial surface and appearance.
Choi et al.~\shortcite{Choi2022} replaced the muscle-based parameterization used earlier~\cite{Sifakis05,srinivasan2021learning} by a collection of muscle fiber curves, whose contraction and relaxation provide a fine-grained parameterization of human facial expression. The approach strikes a balance between the requirements for anatomically-based and artist-friendly models, but comes at the expense of reduced physical accuracy, as the simulation is solely employed in a pre-processing step to acquire an approximate deformation model of muscle fibers.

Similarly, our approach is designed for ease of usability, as it shares the generality and controllability characteristics found in existing linear blendshape models, yet it offers the additional advantage of achieving more physically accurate facial deformation.

\subsection{Morphable 3D Face Models}

As our generalized physical face model is driven by identity and expression and produces a deformed 3D face as the output, it is akin to traditional 3D morphable face models in {current literature~\cite{Blanz1999,FLAME,LYHM,FaceScape,REALY}}, with the main difference that ours allows to simulate physical effects.  A common application of most 3DMMs is their use in monocular face reconstruction, \eg fitting the model to images.  As such, we present a high-level summary of existing techniques where a morphable model is used to recover a person's facial geometry in 3D either by directly optimizing the face shape based on an observed image or through an inference-driven approach that trains neural networks to predict the parameters of a morphable model.

Determining the optimal 3DMM shape, expression, and pose parameters for a given RGB image is achieved through either analysis-by-synthesis optimization~\cite{Gecer_2019_CVPR,gecer2021fast} or deep neural network regression~\cite{Feng:SIGGRAPH:2021, Zielonka2022TowardsMR, Zhang_2023_ICCV}. Recently, there have also been several approaches that rely on additional perception based loss terms to improve the visual quality of these morphable model fits~\cite{EMOCA:CVPR:2021, filntisis2022visual,otto2023perceptual}.

Comprehensive surveys on 3DMMs and their application in monocular face capture are provided by Egger et al.~\shortcite{Egger20}, and Morales et al.~\shortcite{Morales2020SurveyO3}.  As an application of our generalized physical face model we also show the ability to fit the model to unseen identities in the form of a single face image.  However, we do not propose a competing method for accurate 3D geometry reconstruction but rather a convenient approach to obtain an animatable physical model, using a similar fitting approach as in the field of monocular face capture.

\section{Preliminaries: Actuated Face Simulation}
\label{sec:prelim}

In continuum mechanics, motion is characterized by an invertible map $\phi : \X \in \Omega^0 \rightarrow \x \in \Omega $ from the undeformed material space $\Omega^0$ to the deformed space $\Omega$.
The deformation gradient, $\F(\X) = \nabla_{\X} \phi(\X)$, encodes the local transformations including rotation and stretch. The quasi-static state of $\phi$ in the absence of external force is governed by the point-wise equilibrium:
\begin{equation}
   \divergence \P = \divergence \frac{\partial \Psi}{\partial \F}(\F) = \0,
   \label{eqn:governing}
\end{equation}
where 
$\P$ is the first Piola–Kirchhoff stress tensor that measures the internal force. For hyperelastic material, $\P$ is associated with a specific energy density function $\Psi$ that describes the material behavior. Intuitively, \Eqn{governing} means the net force within the material is zero everywhere.

In the context of actuated face simulation, the material space $\Omega^0$ is defined as the undeformed soft tissue space confined between the rest bones $\bonespace^0$ and skin $\skinspace^0$. $\bonespace^0$ consists of the skull $\skullspace^0$ and the jaw  $\jawspace^0$, which will constrain and drag the soft tissue during articulation. The deformed space $\Omega$ is the soft tissue space of the target expression. For $\Psi$, the shape targeting model~\cite{klar2020Shapetargeting} is employed:
\begin{equation}
    \label{eqn:shapetarget}
    \Psi(\F, \A) = \underset{\R \in \SO(3)}{\operatorname{min}}||\F - \R\A||^2_F = ||\F - \R^*\A||^2_F,
\end{equation}
where $\A$ is a symmetric actuation tensor mimicking the local muscle actuation \review{at a single point. Note that when we later refer to a continuous tensor field over the entire space, we will use the notation $\A(\cdot)$}. $\R^*$ is the polar decomposition of $\F\A$, making $\Psi$ rotationally-invariant. 
Based on embedded simulation, $\Omega^0$ is uniformly discretized into a simulation mesh $\simmesh$ using regular elements with nodal vertices $\u^0$, where the discretized skin, skull and jaw are linearly embedded with barycentric weights \review{$\W_{\text{skin}} \u^0$, $\W_{\text{skull}} \u^0$, and $\W_{\text{jaw}} \u^0$ }respectively. With the Finite Element Method (FEM) applied to \Eqn{governing}, the simulation 
then reduces to an energy minimization problem w.r.t. the deformed vertices $\u$ such that the boundary conditions from the articulated bone are satisfied, as follows:
\begin{align}
    &\underset{\u}{\argmin} \sum_{e} \frac{ V_e}{2}|| {\F}_e(\u, \u^0)-\R_e^* \A_e ||_{F}^{2} \\
    &\st \left[
        \begin{array}{l} 
            \W_{\text{skull}} \\
            \W_{\text{jaw}} \\
        \end{array} 
    \right] \u = \left[
        \begin{array}{l} 
            \W_{\text{skull}} \\
            \R_{\text{jaw}} \W_{\text{jaw}} \\
        \end{array} 
    \right] \u^0+ \left[
        \begin{array}{l} 
            \0 \\
            \t_{\text{jaw}} \\
        \end{array} 
    \right],
\end{align}
where $V_e$ is the volume for each element $e$ while $\{\R_{\text{jaw}}, \t_{\text{jaw}}\}$ denotes the rigid transformation for the jaw.

In summary, the simulation entails three core elements:
\begin{itemize}
\item Identity material space ($\Omega^0$) to get $\simmesh$ with embedded skin, skull, and jaw.
\item Facial actuation defined by the tensor field $\act$ over $\simmesh$.
\item Jaw kinematics via mandible transformation $\{\R_{\text{jaw}}, \t_{\text{jaw}}\}$.
\end{itemize}

The latter two components function as muscle actuation mechanisms and are utilized as the input physical constraints. This simulation process can be made efficiently differentiable with the adjoint method, allowing inverse design of $\act$ and  $\{\R_{\text{jaw}}, \t_{\text{jaw}}\}$ from a target expression of the identity. However, the expensive computation from the \review{FEM simulation} and the requirement of the pre-defined material space $\Omega^0$ prevent this method from lending itself to a large face dataset of hundreds of identities, where the identity-specific material space (underlying anatomy) is typically missing.

\section{Method}
\label{sec:Method}

\begin{figure*}
    \centering
    \includegraphics[width=1.\textwidth]{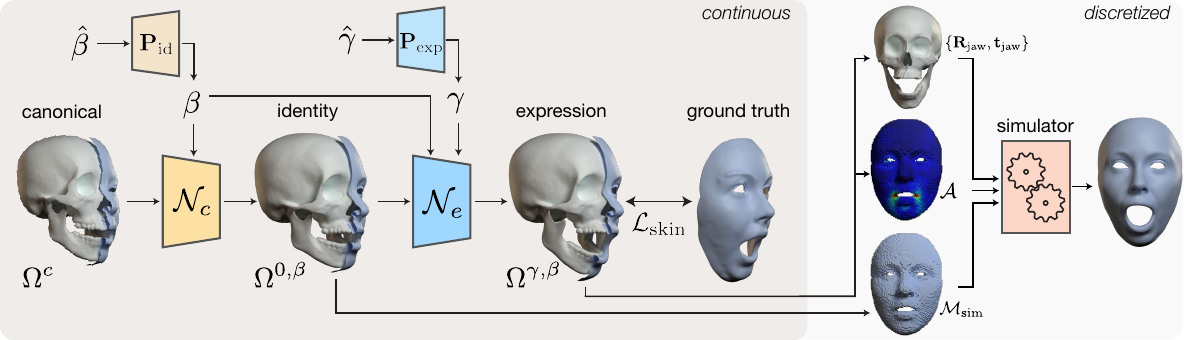}
    \caption{Overview of our model.  Driven by identity $\beta$ and expression $\gamma$ latent codes, $\network_C$ learns to deform a canonical material space $\Omega^C$ to be identity-specific, $\Omega^{0,\beta}$.  Then $\network_e$ learns to deform the material space to match a given expression, $\Omega^{\gamma,\beta}$.  The latent codes are parameterized by a common 3DMM ($\hat{\beta}$, $\hat{\gamma}$).  The network is trained with physically-inspired constraints so that, after discretization, a simulator will produce physics-based facial animation.}
    \label{fig:overview}
\end{figure*}

Our goal is to build a \emph{generalized physical} face model from a large 3D face dataset where only skin geometry is given. This face model should be generative and animatable, such that it can be fit to a variety of input data (\eg 3D scans or face images) and then animated with intuitive controls, providing a convenient mechanism to obtain a physical face model for animation purposes.

To achieve this goal, we propose a new network architecture for implicit actuation mechanisms (illustrated in \Fig{overview}).  At a high level, our model is driven by two latent variables $\beta$ and $\gamma$, which represent the identity and expression, respectively. The output consists of three essential components: the discretized simulation mesh $\simmesh$ at rest, derived from the identity-specific material space $\mathcal{D}^{0, \beta}$; the jaw transformation $\{\R_{\text{jaw}}, \t_{\text{jaw}}\}$; and the actuation tensor field $\act$, both of which are extracted from the corresponding deformed space $\mathcal{D}^{\gamma, \beta}$. These outputs—$\simmesh$, $\{\R_{\text{jaw}}, \t_{\text{jaw}}\}$, and $\act$—are then employed in the FEM simulation to accurately deform the face to perform the given expression.

As mentioned in \Sec{prelim}, actuated face simulation traditionally cannot be trained on large datasets.  Our method is possible thanks to two novel contributions.  The first one is designed for efficient training, where we parameterize and learn the physical constraints in a manner \review{free of any FEM simulation} (\Sec{simulationfree}).
Second, we introduce a material-space generative network that produces the identity-specific material spaces automatically, with no manual modeling (\Sec{ms_generation}).  We will first describe these two contributions in isolation, and then elaborate on how they are combined to provide the complete pipeline (\Sec{phyfacemodel}). \review{Note that throughout the manuscript, we use the term ``simulation-free'' to refer to any process that does not involve FEM simulation.}

\subsection{Simulation-Free Learning}
\label{sec:simulationfree}

Given a material space $\Omega^0$, where the skin $\skinspace^0$, skull $\skullspace^0$ and jaw $\jawspace$ are explicitly defined, we want to infer a continuous actuation field $\A(\cdot)$ and a jaw transformation $\{\R_{\text{jaw}}, \t_{\text{jaw}}\}$ to match a given expression in a simulation-free manner. 
Formally, the objective for the inverse design is as follows:
\begin{align}
    \quad &\quad \quad \quad \underset{\phi(\cdot), \A(\cdot), \R_{\text{jaw}}, \t_{\text{jaw}}}{\argmin} \int_{\skinspace^0} || \phi(\X) - \hat{\phi}(\X) ||^2_2 \, d\X \\
    \label{eqn:continuous_softc}
    \st &\divergence (\nabla_\X\phi(\X) - \R^*(\X)\A(\X)) = \0, \quad \forall \X \in \Omega^0 \; \review{\A = \A^\top}\\
    \label{eqn:continuous_skullc}
    &  \phi(\X) = \X, \quad \forall \X \in \skullspace^0 \\
    \label{eqn:continuous_jawc}
    &\phi(\X) = \R_{\text{jaw}} \X + \t_{\text{jaw}}, \quad \forall \X \in \jawspace^0 \; \R_{\text{jaw}} \in \SO(3) \\
    \label{eqn:realistic}
    &\phi \in \Phi_{\text{bio}} 
\end{align}
where $\hat{\phi}$ denotes the ground truth deformation that is only defined on $\skinspace^0$, \ie through 3D scanning. \Eqn{continuous_softc} comes from the point-wise equilibrium of \Eqn{governing} 
with $\P$ instantiated from \Eqn{shapetarget}. \Eqn{continuous_skullc} and \Eqn{continuous_jawc} guarantee that the skull is fixed and jaw is rigidly articulated. \Eqn{realistic} constrains the mapping $\phi$ such that it resembles bio-mechanically plausible soft tissue deformation, which we refer to as the space $\Phi_{\text{bio}}$ (and will elaborate on later in the discussion of the {\em Soft Loss}).
The motivation of our solution lies in the fact that given any invertible mapping function ${\phi}$, \Eqn{continuous_softc} can be satisfied by setting the actuation tensor field $\A(\cdot)$ as:
\begin{equation}
    \label{eqn:motivation}
    \A(\X) = {\R}_{\phi}(\X)^{\top} \nabla_\X {\phi}(\X),
\end{equation}
where ${\R}_{{\phi}}(\X)$ is the polar decomposition of $\nabla_\X {\phi}(\X)$ at $\X$. This representation gives the zero stress tensor $\P$ and hence the zero divergence everywhere. 
Based on this observation, the inverse design can be simplified to \emph{finding ${\phi}^*$ that closely approximates $\hat{\phi}$ on $\skinspace$ while at the same time satisfying \Eqn{continuous_skullc}, \Eqn{continuous_jawc} and \Eqn{realistic} as much as possible}.
Then, we can compute $\A(\cdot)$ and $\{\R_{\text{jaw}}, \t_{\text{jaw}}\}$ from ${\phi}^*$ automatically. Specifically, $\A(\cdot)$ is obtained with \Eqn{motivation}. $\{\R_{\text{jaw}}, \t_{\text{jaw}}\}$ is obtained with Procrustes alignment between ${\phi}^*(\jawspace^0)$ and $\jawspace^0$. These physical constraints will be compatible with the simulation after discretization with FEM.

To achieve this goal, we parameterize the mapping function $\phi$ with an implicit neutral network $\network_e$. Then, we introduce several novel loss functions to constrain it.

\paragraph{Reconstruction Loss $\loss_\text{skin}$.} The first loss is the reconstruction loss defined on $\skinspace^0$:
\begin{equation}
    \label{eqn:reconstruction}
    \loss_\text{skin}(\skinspace^0) = \sum_{i=1}^{N_v} \frac{1}{N_v}||\network_e(\X_i) - \hat{\x}_i||^2_2,
\end{equation}
where \( \X_i \) represents the \( i \)-th sampled point from \( \skinspace^0 \), and $\hat{\x}_i$ indicates the corresponding ground truth position. \review{We sample \( N_v \) points in total.}
 The correspondence is typically approximated through surface registration.

\paragraph{Rigidity Loss $\loss_\text{rigid}$.} The second loss is to enforce the rigidity of the bone, inspired by \Eqn{continuous_jawc}:
\begin{equation}
    \label{eqn:rigid}
    \loss_\text{rigid}(\partial \Omega_\text{b}^0) = \underset{\R \in \SO(3), \t \in \R^3}{\min} {\sum_{i}^{N_b}} \frac{1}{N_b}||\network_e(\X_i) - (\R{\X}_i + \t)||^2_2,
\end{equation}
where we sample $N_b$ points in total for each region $ \partial \Omega_\text{b}^0$. We apply this loss separately to the skull $\skullspace^0$ and the jaw $\jawspace^0$. Therefore, we have $\loss_\text{rigid} = \loss_\text{rigid}(\skullspace^0) + \loss_\text{rigid}(\jawspace^0)$.

\paragraph{Fixation Loss $\loss_\text{fix}$.} Specifically to $\skullspace^0$, we introduce the third loss to enforce the fixation of the skull area, adapted from \Eqn{continuous_skullc}: 
\begin{equation}
    \label{eqn:fix}
    \loss_\text{fix}(\skullspace^0) = {\sum_{i}^{N_f}} \frac{1}{N_f}||\network_e(\X_i) - \X_i||^2_2,
\end{equation}
where we sample $N_f$ points in total on the skull area $\skullspace^0$. 

\paragraph{Soft Loss $\loss_\text{soft}$.} The fourth loss is to learn bio-mechanically plausible deformation based on the Young's Modulus (\(E\)) and Poisson's Ratio (\(\nu\)), inspired by \Eqn{realistic}. This loss consists of two terms, an elastic one and a volume-preserving one:
\begin{equation}
    \label{eqn:soft}
    \begin{aligned}
        \loss_\text{soft}(\Omega^0) = \sum_{i}^{N_s} \frac{1}{N_s} \min_{\R \in \SO(3)} 
        &\mu ||\nabla_\X \network_{\text{e}}(\X_i) - \R||^2_2 + \\
        \min_{\det(\D) = 1}&\lambda ||\nabla_\X \network_{\text{e}}(\X_i) - \D||^2_2,
    \end{aligned}
\end{equation}
where we sample $N_s$ points in total \review{inside} the material space $\Omega^0$. $\mu$ and $\lambda$ are the Lam\'e parameters, describing the material behavior. These two parameters are parameterized by \(E\) and \(\nu\) as 
$\lambda = E \nu / (1+\nu)(1-2\nu)$ and 
$\mu = E / 2(1+\nu)$ respectively. \review{Consistent with established practices in Projective Dynamics~\cite{Bouaziz2014ProjectiveSimulation}, $\D$ is defined as a matrix with a determinant of 1, matching the dimensions of $\F$.} 
This loss is essential as it not only regularizes the deformation but also implicitly \review{connects the skin and the bone via the soft tissue in-between, ensuring that the deformation of one directly influences the other.} In this regard, when the output skin is supervised towards the ground truth, the jaw is also placed in a constrained position, hence inferring the jaw kinematics. \review{At a high level, the idea of $\loss_\text{soft}$ is to replace the FEM simulation with a neural surrogate that is smoother and more easily differentiable, akin to physics-informed neural networks (PINNs)~\cite{raissi2019physics}.}

These four loss terms ($\loss_\text{skin}$, $\loss_\text{rigid}$, $\loss_\text{fix}$, $\loss_\text{soft}$) will be used in our end-to-end training function defined in \Sec{phyfacemodel}.  The significant benefit of this formulation is that we move the traditional FEM-discretized simulation out of the learning loop, which makes the method much faster and easily scalable since all the loss functions are point-wise. 
In summary, this network generates physically-constrained deformations that are strategically converted into physical constraints used in the simulation.

\subsection{Material Space Morphing}
\label{sec:ms_generation}

We now describe our second main contribution to address the fact that every identity in the dataset needs a custom material space.  Our approach is to infer the material space of an identity automatically by learning to morph a single canonical material space to any new person, \review{similar to the idea of canonical warping \cite{yang2023ExprAndStyle,zheng2022imface,kirschstein2023nersemble}.}

Given a canonical material space $\Omega^c$, where the canonical skin $\skinspace^{c}$ and bones $\bonespace^c$ are explicitly defined, we want to morph it into the material space $\Omega^0$ of an identity using another implicit network $\network_{c}$. We propose three loss functions to constrain the solution during training.

\paragraph{Identity Loss $\loss_\text{id}$.} The first loss is to provide supervision on the skin area, similar to \Eqn{reconstruction}.
\begin{equation}
    \label{eqn:id_reconstruction}
    \loss_\text{id}(\skinspace^c) = \sum_{i=1}^{N_v} \frac{1}{N_v}||\network_{c}(\X^c_i) - \hat{\X}_i||^2_2,
\end{equation}
where \( \X^c_i \) represents the \( i \)-th sampled point from \( \skinspace^c \), and $\hat{\X}_i$ indicates the corresponding ground truth position on the neutral identity mesh in the dataset, attainable from 3D scanning. 

\paragraph{Bone Shape Loss $\loss_\text{bone}$.} Since we only have direct supervision on the skin area, we propose to constrain the bone shapes using an off-the-shelf parametric bone generator~\cite{qiu2022sculptor}, which can predict plausible skull and jaw shapes given a neutral face mesh.  Formally, the loss is as follows:

\begin{equation}
    \label{eqn:boneshape}
    \loss_\text{bone}(\bonespace^c) = \sum_{i=1}^{N_{b}} \frac{1}{N_{b}}||\network_{c}(\X^c_i) - \hat{\X}_i||^2_2,
\end{equation}
where we sample $N_{b}$ points on $\bonespace^c$ in total, and here $\hat{\X}_i$ is the pseudo-ground truth bone position on the bone surfaces generated by the parametric model given the ground truth identity neutral shape.

\paragraph{Elastic Regularization $\loss_\text{ereg}$.} 
Finally, since we only supervise on the surface areas, we incorporate an elastic regularization to smooth the volumetric morphing:
\begin{equation}
    \loss_\text{ereg}(\Omega^c) = \sum_{i}^{N_s} \frac{1}{N_s} \min_{\R \in \SO(3)} 
    ||\nabla_{\X^c} \network_c(\X^c_i) - \R||^2_2, 
    \label{eqn:elastic}
\end{equation}
where we sample $N_{s}$ points on $\Omega^c$ in total. This loss also makes the training robust to any incorrect estimation from the bone prediction in \Eqn{boneshape}.
These three loss terms ($\loss_\text{id}$, $\loss_\text{bone}$, $\loss_\text{ereg}$) will be added to our end-to-end training approach, described next.

\subsection{Generalized Physical Face Model}
\label{sec:phyfacemodel}

We can now describe our complete pipeline, illustrated in \Fig{overview}. The two contributions described in \Sec{simulationfree} and \Sec{ms_generation} make it possible to learn a generalized physical face model from a large dataset of 3D facial scans (skin only).  To summarize, given an identity latent code $\beta$, our generative implicit network $\network_c$ learns to deform points from a canonical material space $\Omega^c$ to an identity-specific material space $\Omega^{0, \beta}$.  Then given both the identity code and an expression latent code $\gamma$, our implicit deformation network $\network_e$ generates the identity- and expression- specific deformations $\Omega^{\gamma, \beta}$, from which the physical constraints can be obtained for simulation.

In order to regularize the identity and expression latent spaces, encourage disentanglement, and allow for artist-driven animation after training, we parameterize the latent codes using an off-the-shelf morphable 3D face model \cite{FLAME}.  Specifically, $\beta = \mathbf{P}_{\text{id}}(\hat{\beta})$ and $\gamma = \mathbf{P}_{\text{exp}}(\hat{\gamma})$, where $\mathbf{P}_{\text{id}}$ and $\mathbf{P}_\text{exp}$ are small MLPs with three layers each that \review{operate like learnable position encoding}, and $\hat{\beta}$ and $\hat{\gamma}$ are the identity and expression parameters of the 3DMM. To obtain the input for training, we pre-compute the 3DMM parameters corresponding to each face in our dataset using least-squares fitting.

\paragraph{Complete Loss Function.} 
Our full set of optimization variables include the network weights of $\network_c$, $\network_e$, $\mathbf{P}_\text{id}$ and $\mathbf{P}_\text{exp}$.  We regularize the latent codes with $l$-2 regularization $\loss_{\text{lreg}}=||\beta||_2^2 + ||\gamma||^2_2$.  For smoothness, we also apply Lipschitz regularization $\loss_{\text{lip}}$ to $\mathbf{P}_\text{id}$ and $\mathbf{P}_\text{exp}$, as in Yang et al.~\shortcite{yang2023ExprAndStyle}. Putting it all together, the complete objective function for training our face model is:

\begin{equation}
    \label{eqn:training}
    \begin{aligned}
        \loss_\text{train} &= \lambda_\text{skin}\loss_{\text{skin}} + \lambda_\text{rigid}\loss_{\text{rigid}} +  \lambda_\text{fix}\loss_{\text{fix}} +  \lambda_\text{soft}\loss_{\text{soft}} \\
        &+ \lambda_{\text{id}}\loss_{\text{id}} + \lambda_{\text{bone}}\loss_{\text{bone}} + 
        \lambda_{\text{ereg}}\loss_{\text{ereg}} \\
        &+ \lambda_\text{lreg}\loss_{\text{lreg}} + \lambda_\text{lip}\loss_{\text{lip}},
    \end{aligned}
\end{equation}
where $\lambda_{*}$ are balancing weights. Our model is trained end-to-end in a simulation-free manner with the direct supervision only from the skin scans while the anatomical features such as the bone shapes, jaw kinematics, and the facial actuation, are inferred automatically. 

\paragraph{Test-time Optimization and Applications.} 
Once trained, one application of our model is to fit it to unseen identities. To accomplish this, we optimize the latent codes $\beta$ and $\gamma$ using an objective \review{that mirrors the simulation-free approach of our training phase}:

\begin{equation}
    \label{eqn:testing}
    \begin{aligned}
        \loss_\text{test} &= \lambda_\text{skin}\loss_{\text{skin}} + \lambda_\text{rigid}\loss_{\text{rigid}} +  \lambda_\text{fix}\loss_{\text{fix}} +  \lambda_\text{soft}\loss_{\text{soft}} \\
        & + \lambda_{\text{bone}}\loss_{\text{bone}} + 
        \lambda_{\text{ereg}}\loss_{\text{ereg}} + \lambda_\text{lreg}\loss_{\text{lreg}},
    \end{aligned}
\end{equation}
\review{where we exclude $\loss_{\text{id}}$, eliminating the need for the input to depict a neutral expression, as well as the irrelevant $\loss_{\text{lip}}$ that is intended for regularizing network weights.}
During fitting, $\loss_{\text{skin}}$ and $\loss_{\text{bone}}$ can vary depending on the form of the input data. For example, when fitting to an unseen 3D scan we can use the same scan-to-mesh loss and predicted bone loss (\ie \Eqn{reconstruction} and \Eqn{boneshape}) as during training.  However, our model is flexible and allows fitting to other data, such as 2D facial landmarks computed from an image, where we can formulate $\loss_{\text{skin}}$ as a 2D projection loss and $\loss_{\text{bone}}$ as a self-supervised bone loss, as we will describe later in \Sec{fitting}. 
\review{Notably, for differing head poses in world-space, optimization includes the rigid skull transformation.}

\review{Once the latent codes are optimized, we can generate the material space via $\network_c$ and evaluate the physical constraints with $\network_e$. After discretization, we obtain the simulation mesh $\simmesh$, the actuation tensor field $\mathcal{A}$ over $\simmesh$, and the jaw transformation $\{\R_{\text{jaw}}, \t_{\text{jaw}}\}$, which can be used for the FEM simulation, as described in \Sec{prelim}. With an appropriate discretization resolution, the difference between the fitted result and the simulated result is negligible, as we show in \Sec{evaluation}.}

Finally, we can directly manipulate the physical model from the artist-friendly 3DMM parameters, or swap the identity latent code for animation retargeting.  We can also interpolate between identity codes for novel face generation.  
The main superiority of our physical face model over other traditional deep face models is that during simulation, we can add additional physical effects such as collision handling, external forces, etc.  All of these applications will be demonstrated in \Sec{Results}.

\section{Implementation}
\label{sec:Implementation}

In order to train our network there are a number of implementation details to consider.

\paragraph{Training Data.}
We use the 3D face dataset presented by Chandran et al.~\shortcite{chandran2020semanticface}, which consists of 336 identities under various expressions totaling 13000 face scans in topological correspondence, with rigid head motion removed.  Every identity has one ``neutral'' expression that is used in \Eqn{id_reconstruction}.  
As the dataset contains only skin geometry, we pre-fit the bones using a parametric skeletal model called SCULPTOR~\cite{qiu2022sculptor}, which provides the constraints for the bone shape loss in \Eqn{boneshape}. 
The skin and the bone surfaces discretize $\skinspace^*$ and $\bonespace^*$, respectively. As mentioned in \Sec{phyfacemodel} of the main text, 
we fit FLAME model parameters ($\hat{\beta}$, $\hat{\gamma}$) to all data in a pre-processing step.  

\paragraph{Canonical Space.}
The canonical material space $\Omega^c$ is defined using the mean bone and skin surfaces of SCULPTOR.  For the bones we use the topology directly from SCULPTOR, but for the facial skin we fit the topology from our 3D face dataset to the SCULPTOR mean face in order to obtain vertex consistency with our dataset, for the constraints in \Eqn{reconstruction} 
and \Eqn{id_reconstruction}.

\paragraph{Training Details.}
All the losses defined on $\skinspace^*$ $\bonespace^*$, $\skullspace^*$, and $\jawspace^*$ are evaluated directly on the mesh vertices, while losses defined in $\Omega^*$ are computed through uniform sampling of the soft tissue space. 
We mask out the necessary face and bone regions for regression. 
Specifically, we only consider the frontal face and assign 10 times less supervision weight to the low-confidence regions (confidence per vertex is available in the 3D dataset). During testing and evaluation, we use the same masks for consistency. \review{We train the model for 200 epochs using the Adam optimizer \cite{kingma2015adam} with a batch size of 16 and an initial learning rate of $1e-4$ that remains the same for the first 100 epochs and starts to linearly decay to zero after.}

\paragraph{Hyper-Parameters.}
The balancing weights for training are set as follows: $\lambda_\text{skin}=20$, $\lambda_\text{rigid}=20$, $\lambda_\text{fix}=20$, $\lambda_\text{soft}=0.1$, $\lambda_\text{id}=1$, $\lambda_\text{bone}=0.1$, $\lambda_\text{lreg}=1e-4$, $\lambda_\text{ereg}=0.1$, and $\lambda_\text{lip}=2e-6$.
The number of sampling points are \( N_v = 45k \), \( N_b = 5k \), \( N_f = 10k \) and \( N_s = 10k \). The values for density, Young's Modulus (\( E \)) and Poisson ratio (\( \nu \)) are set to 0.9 g/ml, 5 kPa and 0.47, respectively, according to related literature~\cite{xu2015human}.

\begin{figure}
    \centering
    \includegraphics[width=0.5\textwidth]{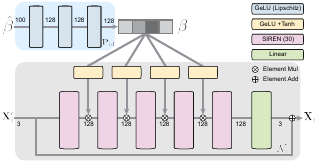}
    \caption{\review{Detailed architecture of the identity branch. The size of the inputs, the latent codes, and the output are shown on the links.}}
    \label{fig:network}
\end{figure}

\review{\paragraph{Network Architecture.}
The dimensions of our latent codes $\beta$ and $\gamma$ are set to 128. For parameterization $\hat{\beta}$ we use the first 100 principal components of the identity space of the FLAME model~\cite{FLAME}, while $\hat{\gamma}$ consists of the full expression space as well as the 3-dimensional joint pose.
The parameterization MLPs for $\P_{\text{id}}$ and $\P_{\text{exp}}$ consist of 3 fully connected GeLU layers~\cite{hendrycks2016gaussian} each with 128 hidden units and the Lipschitz weight normalization~\cite{liu2022learning}. 
For both $\network_{e}$ and $\network_{c}$, we use a conditional SIREN network inspired by Yang et al.~\shortcite{yang2022implicit} for its sound differential properties, which are essential for evaluating the physical constraints.
The backbone consists of 5 SIREN layers~\cite{siren2020} each with 128 hidden units and frequency 30, followed by a linear layer to output the displacement that is then added to the input vertices to get the final output. The conditioning is achieved through the weight modulation mechanism proposed by Yang et al.~\shortcite{yang2022implicit}. A detailed architecture for the identity branch is shown in \Fig{network}. The expression branch is similarly structured.
}

\review{\paragraph{Simulation.}
We employ hexahedral elements,  with a simulation mesh example depicted in \Fig{overview}.
The discretization resolution is approximately $2$mm based on the study shown in \Sec{evaluation}.
Once the simulation mesh $\simmesh$ is discretized, we sample the points inside $\simmesh$ to evaluate the actuation tensor field $\act$. For simulation, we use the same solver as Yang et al.~\shortcite{yang2022implicit,yang2023ExprAndStyle}. }

\paragraph{Test Set.}
For testing, we prepare two datasets. The first dataset is a static dataset that contains 28 unseen identities totaling 529 scans, which will be used to examine the model's generalization to unseen identities with diverse expressions. The second dataset is a dynamic dataset consisting of 5 unseen performance sequences of 5 seen identities, with each sequence lasting around 10s. This dynamic test set will be used to examine the model's generalization to unseen expression blendweight vectors. This is valuable for animation retargeting purposes.

\paragraph{Timing.}
The detailed timing results are presented in \Tab{timing}. During the training phase, the average duration per iteration is approximately 0.5 seconds. In the testing phase, for each frame, the latent space optimization (Optim.) requires about 7 seconds, whereas the generation of the input physical constraints (PhyCons.) takes approximately 0.05 seconds, and the simulation process (Sim.) takes around 2.6 seconds. These timing experiments were conducted on a system equipped with a single RTX A6000 GPU and a 16-core CPU.

\begin{table}
    \caption{Timing of different components.}
    \begin{tabularx}{0.48\textwidth}{>{\raggedleft\arraybackslash}X|>{\raggedleft\arraybackslash}X|>{\raggedleft\arraybackslash}X|>{\raggedleft\arraybackslash}X|>{\raggedleft\arraybackslash}X}
        \toprule
        \multicolumn{2}{c|}{\textbf{Training}} & \multicolumn{3}{c}{\textbf{Testing}} \\
        \cline{1-5}
        \customstrut{2.5ex} Total            & Per Iter     & Optim.       & PhyCons.          & Sim. \\
        \hline
        \customstrut{2.5ex} 23.13h           &0.50s         & 6.98s       & 0.05s             & 2.58s       \\ 
        \bottomrule
    \end{tabularx}
    \label{tab:timing}
\end{table}

\section{Results}
\label{sec:Results}

We now demonstrate results and applications of our generalized physical face model, starting with fitting to unseen data (\Sec{fitting}), showcasing the benefit of having a physical model by illustrating physical effects (\Sec{physicalEffects}), facial animation retargeting (\Sec{retargeting}), and identity generation/blending through latent space interpolation (\Sec{interpolation}). Please refer to the supplemental video for a more vivid visualization.

\subsection{Fitting to Unseen Data}
\label{sec:fitting}

The primary benefit of our face model is its ability to fit to unseen identities automatically, alleviating the expensive burden of creating an identity-specific material space by hand and training an identity-specific actuation network.  Fitting proceeds by optimizing for the identity and expression latent codes ($\beta$, $\gamma$) using \Eqn{testing}.  Here, the skin loss $\loss_{\text{skin}}$ and bone loss $\loss_{\text{bone}}$ are adapted based on the type of data being fit.  We illustrate fitting to two different modalities: fitting to a single 3D face scan and fitting to a single face image.

\begin{figure}
	\centering
    \includegraphics[width=1.0\linewidth]{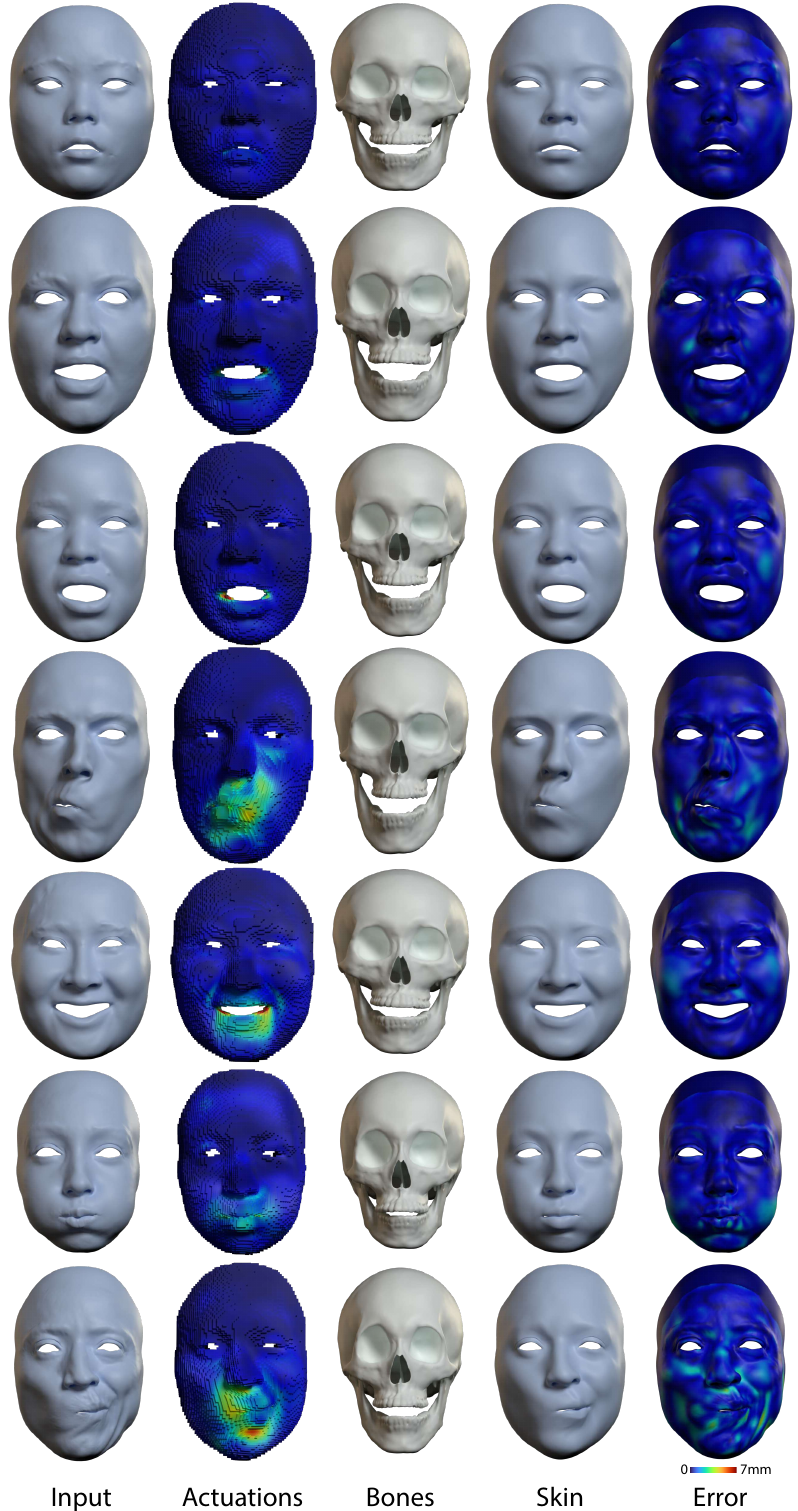}
    \caption{Model fitting to a 3D scan.  \review{Seven} examples are shown for different identities and different expressions.  After fitting, the predicted actuations and bones allow to simulate the final facial skin mesh, which matches the input with a very low error. }
    \label{fig:scan_reconstruction}
\end{figure}

\paragraph{Fitting to a 3D Scan.}
It has become common practice to perform at least a small amount of 3D face scanning for the primary actors in high-end productions.  Given just a single scan, our model can be fit to provide physics-based animation.  When fitting to a 3D scan, $\loss_{\text{skin}}$ and $\loss_{\text{bone}}$ are defined the same way as during training (\Eqn{reconstruction} and \Eqn{boneshape}).  Several fitting results are shown in \Fig{scan_reconstruction}, including the input scan, the predicted actuations (for the predicted simulation meshes), the estimated bone shapes and mandible transformation, and the final simulated facial skin surface.  We highlight the very low error between the simulated result and the input scan.  Note that the input scan does not need to be in a rest configuration since we fit both the identity and the expression parameters.  %

\begin{figure}
	\centering
    \includegraphics[width=1.0\linewidth]{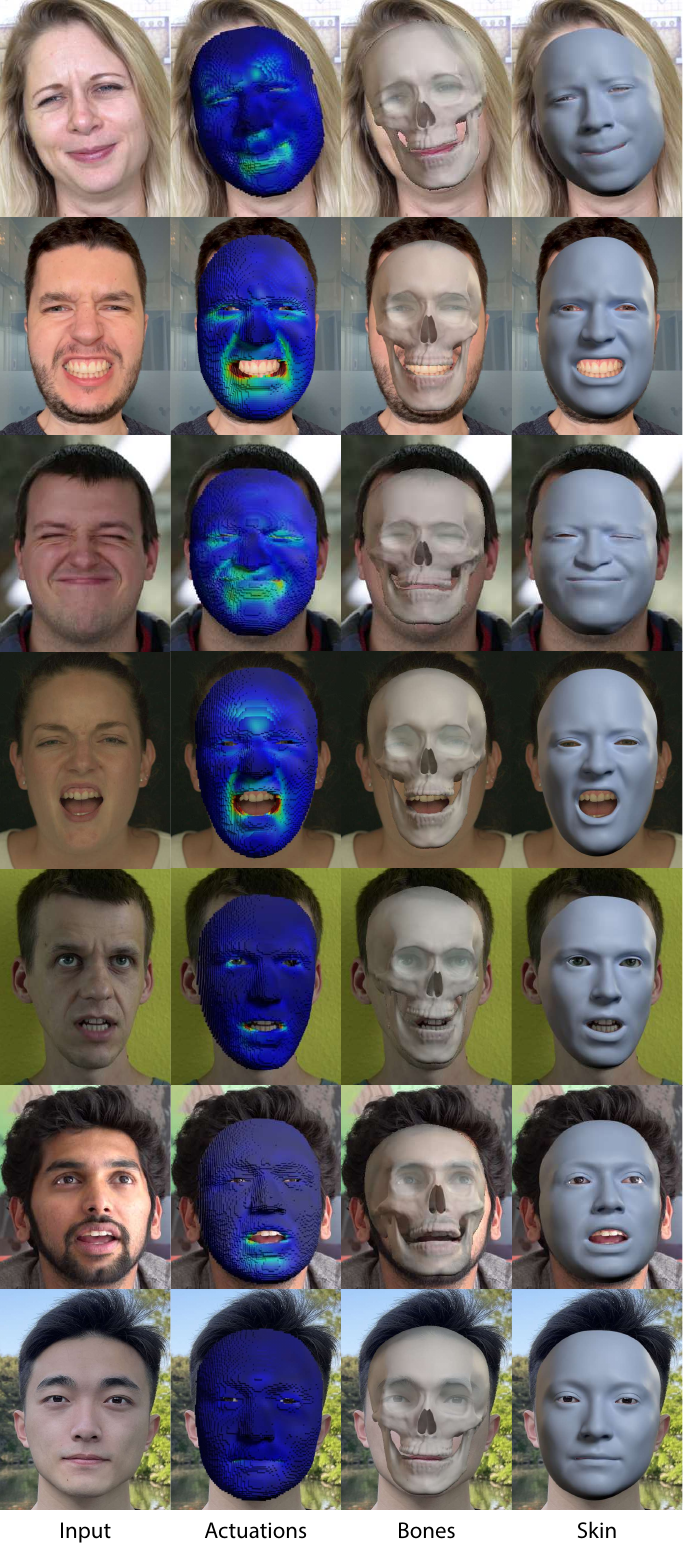}
    \caption{Model fitting to a single face image.  \review{Seven} different identities and expressions are shown.  After fitting, the predicted actuations and bones allow to simulate the final facial skin mesh.  The result is an animatable physical model from a very lightweight input.}
    \label{fig:image_reconstruction}
\end{figure}

\paragraph{Fitting to a Face Image.}
When a 3D scan is not available, our model is flexible and can be fit even to a single face image.  When fitting to an image the skin loss is defined in terms of 2D facial landmarks, and our method predicts not only $\beta$ and $\gamma$ but also the camera projection matrix $\mathbf{C}$.  $\loss_{\text{skin}}$ is then defined as: 

\begin{equation}
    \label{eqn:landmark_reconstruction}
    \loss_\text{skin}(\skinspace^0) = \sum_{i=1}^{N_l} \frac{1}{N_l}||\mathbf{C}\cdot\network_e(\X_i) - \hat{\x}_i||^2_2,
\end{equation}
where $\X_i$ represents the $i$-th landmark out of the $N_l$ landmarks sampled from $\skinspace^0$, and $\hat{\x}_i$ indicates the corresponding ground truth 2D landmark position in screen space.  To obtain the ground truth landmarks we employ a recent high-quality dense landmark detector~\cite{chandran2023continuous}  and predict approximately 8000 landmarks distributed on the face (please refer to the supplemental video for an illustration).  In this scenario, there is no ground truth neutral scan to obtain a predicted bone shape for the bone shape loss $\loss_{\text{bone}}$ in \Eqn{boneshape}. However, we can reformulate $\loss_{\text{bone}}$ as a self-supervised bone loss using the estimated skin to predict the bones:

\begin{equation}
    \label{eqn:boneshape2}
    \loss_\text{bone}(\bonespace^c) = \sum_{i=1}^{N_{b}} \frac{1}{N_{b}}||\network_{c}(\X^c_i) - \mathcal{P}(\network_{c}(\skinspace^c))(\X^c_i)||^2_2,
\end{equation}
where $\mathcal{P}$ represents the parametric bone generator~\cite{qiu2022sculptor}, evaluated on the regressed neutral facial skin $\network_{c}(\skinspace^c)$. 
Fitting results are shown in \Fig{teaser} and \Fig{image_reconstruction}. We systematically evaluate a variety of images with differing resolutions and lighting conditions.  While not as accurate as fitting to a ground truth 3D scan, the image fitting results showcase our ability to preserve critical facial features while providing the most flexible and easy-to-employ version of our model.

For all fitting results, either to a 3D scan or an image, once the latent parameters are determined then the model can be animated as illustrated in \Fig{teaser} and the supplemental video.

\subsection{Physical Effects}
\label{sec:physicalEffects}

Our model allows the simulation of physical effects, which is one of the key benefits of using physically-based animation. 

\begin{figure}
    \centering
    \includegraphics[width=1.0\linewidth]{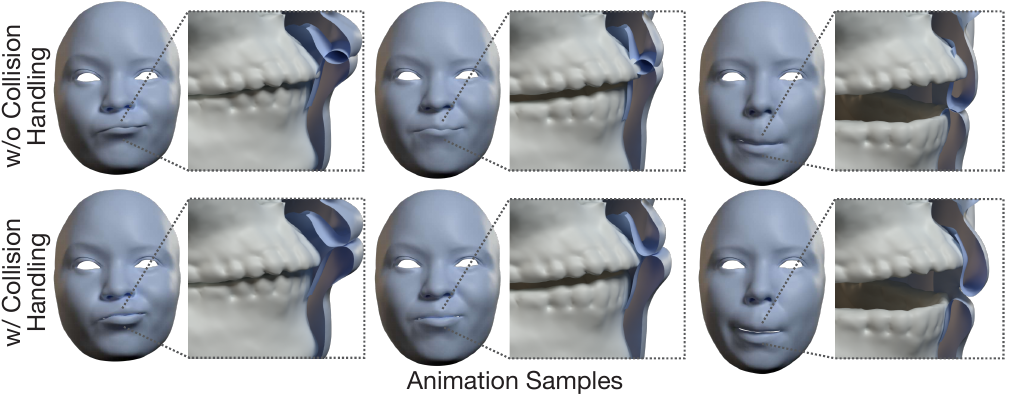}
    \caption{Collision handling. Our physical model is able to accurately detect and resolve interpenetrating geometries.}
    \label{fig:collision}
\end{figure}

\begin{figure}
    \centering
    \includegraphics[width=1.0\linewidth]{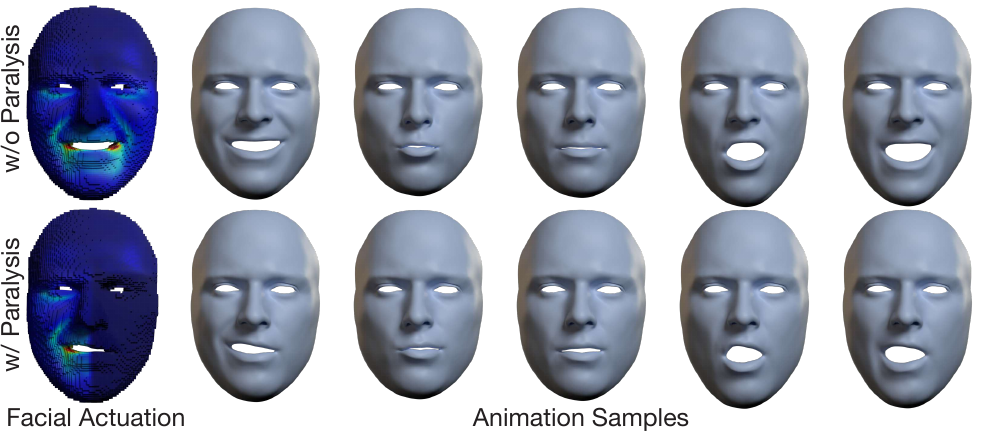}
    \caption{Paralysis. Our physical model can simulate some degrees of facial paralysis, demonstrating the model's sensitivity and precision in depicting subtle physiological changes.}
    \label{fig:paralysis}
\end{figure}

\begin{figure}
    \centering
    \includegraphics[width=1.0\linewidth]{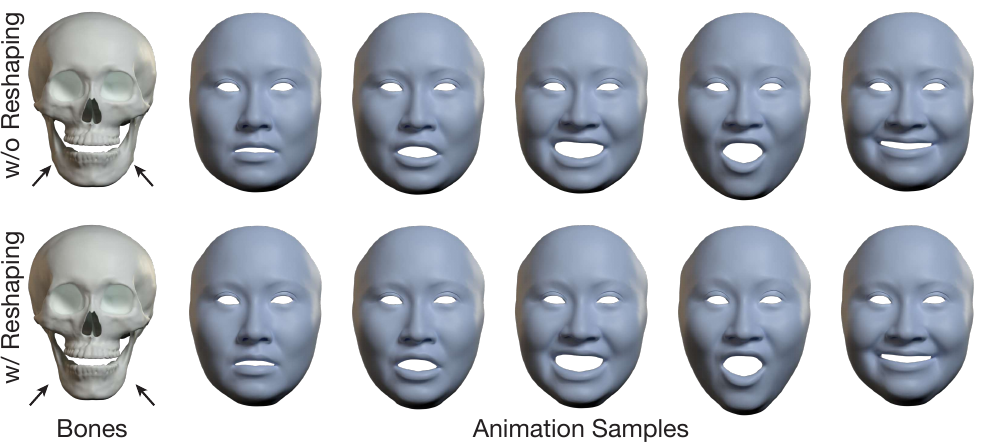}
    \caption{Bone reshaping. Our physical model can simulate various craniofacial effects like osteotomy, where the jaw has been scaled down and the actuations are replayed on the edited anatomy. \review{The top row shows the pre-treatment state, and the bottom row shows the post-treatment state.}}
    \label{fig:osteotomy}
\end{figure}

\begin{figure}
    \centering
    \includegraphics[width=1.0\linewidth]{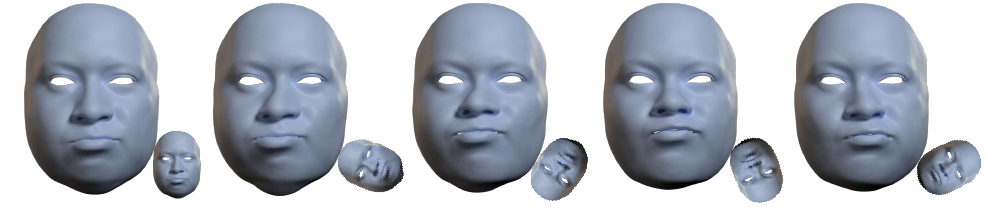}
    \caption{Gravity. We show the effect of gravity on a stabilized face with different head orientations (inset).}
    \label{fig:gravity}
\end{figure}

\paragraph{Collision Handling.}
\Fig{collision} illustrates our model's capacity to accurately detect and resolve collisions, including lip-lip and tooth-lip penetrations, a common challenge in facial animation. \review{The collision resolution is achieved by adding an efficient incremental potential contact energy \cite{li2023efficient} on top of the baseline (\Sec{prelim}).}

\paragraph{Paralysis.}
By adjusting muscle actuation parameters, our model can replicate some degrees of facial paralysis as illustrated in \Fig{paralysis}.  This example demonstrates the model's sensitivity and precision in depicting subtle physiological changes.

\paragraph{Bone Reshaping.}
Our physical model can simulate various craniofacial effects like osteotomy, illustrated in \Fig{osteotomy} where the jaw bone has been scaled down, showing the comparative analysis between pre- and post-treatment states and highlighting our model's effectiveness in representing and adapting to skeletal deformations.

\paragraph{Gravity.}
Our model can simulate the effects of gravity. \Fig{gravity} shows this effect where we rotate the head to different orientations and the soft tissue is naturally pulled in the corresponding direction.

\subsection{Retargeting}
\label{sec:retargeting}
\begin{figure}[t]
    \centering
    \includegraphics[width=1.0\linewidth]{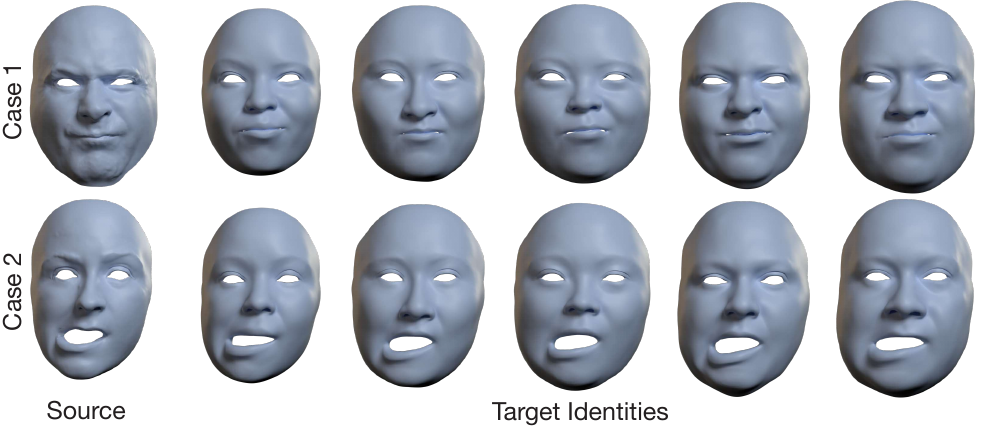}
    \caption{Our physical model is able to transfer animations \review{from a source (left) to multiple target identities (right)} while being interpenetration-free. }
    \label{fig:retargeting}
\end{figure}

Our model can be used for physics-based animation retargeting, where we transfer facial animations between identities without interpenetration (see \Fig{retargeting}). This is accomplished by changing the identity latent code to a target subject while keeping the expression code of the source subject.  This application confirms our model's ability to maintain realism and physical integrity while producing identity-specific facial expressions that match a source input.

\subsection{Latent Space Interpolation}
\label{sec:interpolation}

\begin{figure}[t]
    \centering
    \includegraphics[width=1.0\linewidth]{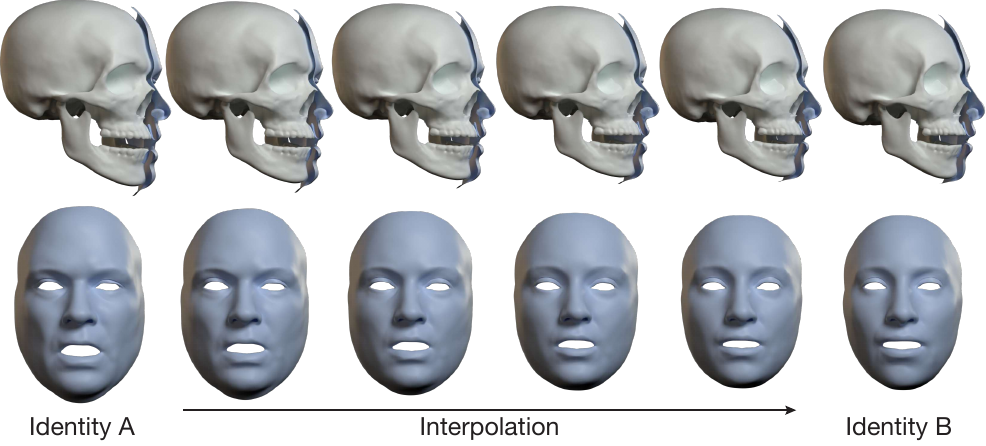}
    \caption{Our physical model is able to smoothly interpolate between different identities (\review{identity codes}) while preserving physical plausibility, such as collision-free properties.}
    \label{fig:interpolation}
\end{figure}

Another application of our generalized model is that we can sample new identities from the latent space.  We demonstrate this by interpolating between two different identities from our training set in \Fig{interpolation}.  For each novel identity, the model can be evaluated with different expression codes to obtain physics-based facial animation.

\section{Evaluation}
\label{sec:evaluation}

We evaluate our model from multiple perspectives numerically, 
including reconstruction accuracy, jaw rigidity, skull fixation and anatomical fidelity.
For reconstruction accuracy, we adopt four types of metrics, including
3D vertex-to-vertex error (V2V), 3D scan-to-mesh error (S2M), F-Score, and normal error.
Finally, we evaluate anatomical fidelity in terms of the bone fidelity and the bone-skin penetration. The specific details of each metric are elaborated as follows:
\begin{itemize}
    \item \textbf{Vertex-to-vertex Error (V2V)} measures the average of the Euclidean distances between the ground-truth 
    and the reconstruction vertices. 
    \item \textbf{Scan-to-mesh Error (S2M)} measures the average of the Euclidean distances between the ground-truth vertices and the reconstructed mesh surface.
    \item \textbf{F-Score} evaluates the reconstruction quality from the point cloud aspect. It harmonizes the recall and the precision by computing their harmonic mean. A high F-Score is indicative of a reconstruction that is both accurate and complete. We sample 32k points in total and use 1mm as the error threshold.
    \item \textbf{Normal Error} measures the average of the cosine distances between the ground-truth and the reconstruction normals.
    \item \textbf{Jaw Rigidity} quantifies how rigidly the network moves the jaw, based on V2V metric (see \Eqn{rigid}). 
    \item \textbf{Skull Fixation} quantifies how well the network fixes the skull, based on V2V metric (see \Eqn{fix}). 
    \item \textbf{Bone Fidelity} quantifies how well the network preserves the SCULPTOR bone space, based on V2V metric (see \Eqn{boneshape}). 
    \item \textbf{Penetration Pairs} counts the penetration between the bone and the skin meshes. We use the edge-triangle pair. 
\end{itemize}

\begin{table}[t]
    \caption{Numerical evaluation of the simulated results across various simulation resolutions and network's output.}
    \begin{tabularx}{0.48\textwidth}{p{1.65cm}|>{\raggedleft\arraybackslash}X|>{\raggedleft\arraybackslash}X|>{\raggedleft\arraybackslash}X|>{\raggedleft\arraybackslash}X}
        \toprule
        \customstrut{2.5ex}\textbf{Res. (mm)} & \textbf{V2V $\downarrow$} & \textbf{S2M $\downarrow$} & \textbf{F-Score $\uparrow$} & \textbf{Normal $\downarrow$} \\ 
        \hline
        \customstrut{2.5ex} $6.8$      &1.4679 &0.6615 &0.7346 &0.0178 \\ 
        \customstrut{2.5ex} $4.5$      &1.1270 &0.5466 &0.8033 &0.0157 \\ 
        \customstrut{2.5ex} $3.0$      &0.9469 &0.4892 &0.8427 &0.0146 \\ 
        \customstrut{2.5ex} \boldsymbol{$2.0$} (Ours)      &0.8975 &0.4721 &0.8546 &0.0143 \\ 
        \customstrut{2.5ex} $1.3$     &0.8848 &0.4666 &0.8583 &0.0142 \\
        \hline
        \customstrut{2.5ex} Network        &0.8642 &0.4532 &0.8672 &0.0141 \\  
        \bottomrule
    \end{tabularx}
    \label{tab:res}
\end{table}

\begin{figure}[t]
    \includegraphics[width=1.0\linewidth]{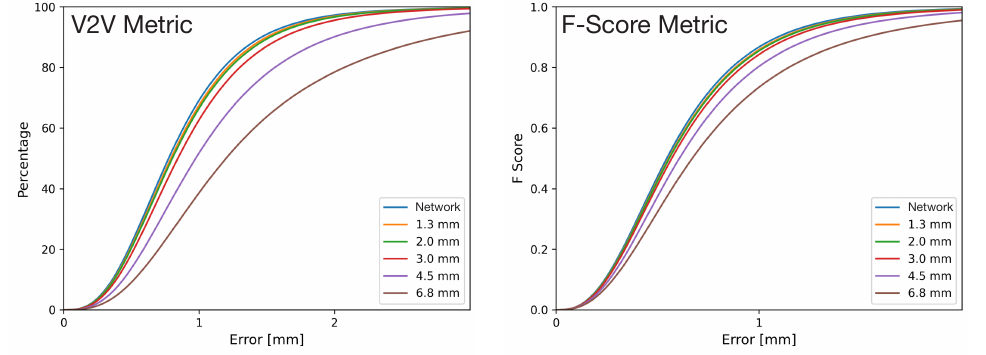}
    \caption{Cumulative curves of V2V and F-Score metrics from the simulated results across various simulation resolutions and network's output.}
    \label{fig:cumulative}
\end{figure}

\begin{figure}
    \includegraphics[width=1.0\linewidth]{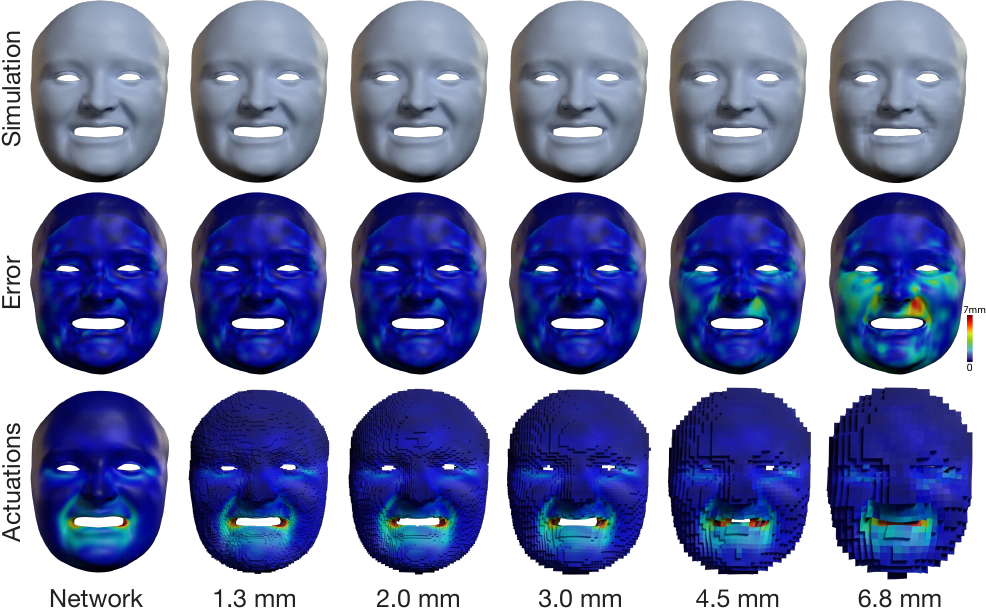}
    \caption{Qualitative comparison of the simulated results across various simulation resolutions and network's output.}
    \label{fig:res}
\end{figure}

\paragraph{Model Accuracy.} 
We first evaluate our model accuracy in terms of fitting accuracy. Specifically, we fit our model to the static test set by optimizing our latent codes scan by scan, using $\loss_\text{test}$ \Eqn{testing}. 
We then run the discretization and the simulation to get the results for calculating the metrics.
\Tab{res} reports the accuracy of our model with different discretization resolutions. The discretization resolution largely impacts the simulation accuracy. The higher the resolution is, the more accurate the simulated results are.  We include a row for evaluating the pure fitted output of the network, which has the lowest error as this is the target of the training objective.  With a discretization resolution of around $2$mm we can achieve comparable accuracy with our network's outputs, substantiating that our {simulation-free} learning framework is effective in inferring plausible physical constraints used in simulation.
\Fig{cumulative} plots the cumulative curves on V2V and F-Score metrics.
\Fig{res} further demonstrates that our simulation results can achieve high reconstruction quality.

To further evaluate our model in terms of animation quality, we test our model on the dynamic set, where we directly use the unseen expression blendweight vectors to animate the seen identities without any optimization. \Tab{ablation} shows that the model generalizes well to the unseen expression blendweights, paving the way for animation retargeting. 
We also report the jaw rigidity and skull fixation in both datasets, proving that the constraints are well enforced (see \Tab{ablation}).

\begin{figure}
    \includegraphics[width=1.0\linewidth]{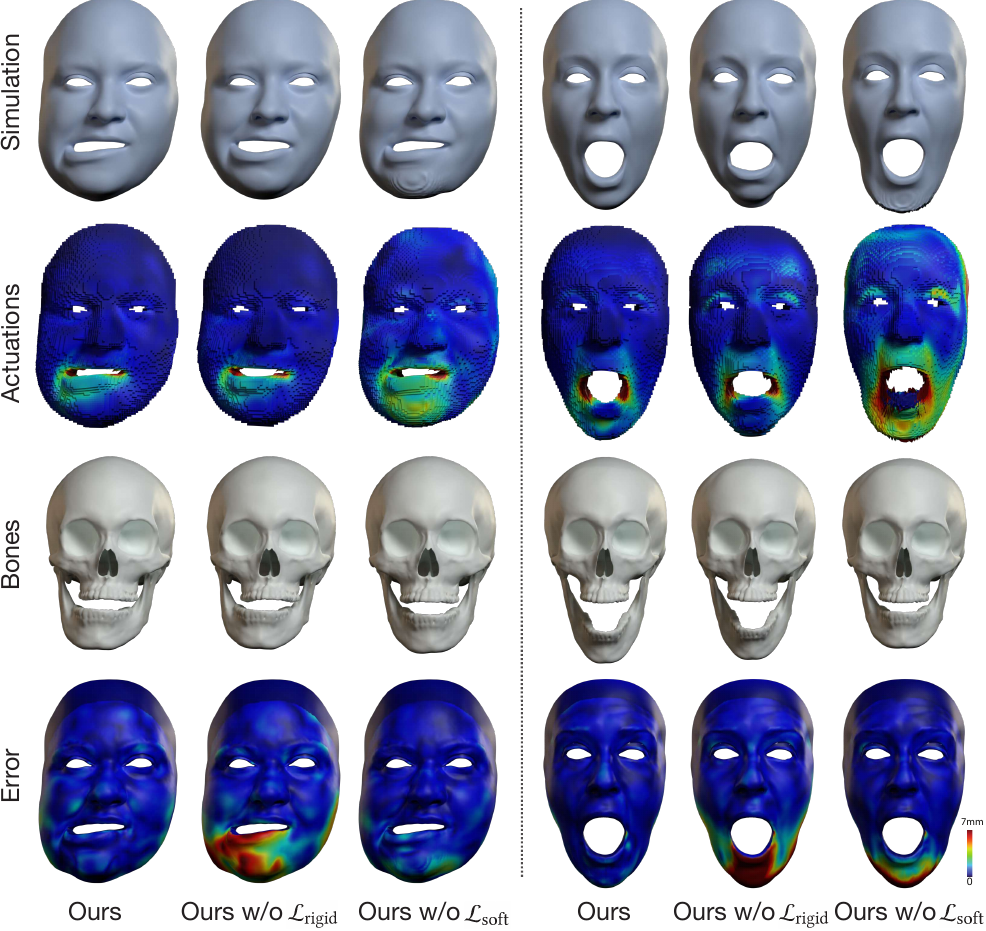}
    \caption{Qualitative comparison of different methods.}
    \label{fig:ablation}
\end{figure}

\begin{table*}
    \caption{Numerical evaluation of different components on static and dynamic test sets. We turn on the latent space optimization for static set while turn off it for dynamic set.}
    \begin{tabularx}{\textwidth}{c|p{3cm}|>{\raggedleft\arraybackslash}X|>{\raggedleft\arraybackslash}X|>{\raggedleft\arraybackslash}X|>{\raggedleft\arraybackslash}X|>{\raggedleft\arraybackslash}X|>{\raggedleft\arraybackslash}X}
        \toprule
        & \textbf{Methods} & \textbf{V2V $\downarrow$} & \textbf{S2M $\downarrow$} & \textbf{F-Score $\uparrow$} & \textbf{Normal $\downarrow$} & \textbf{Jaw Rig. $\downarrow$} & \textbf{Skull Fix.$\downarrow$}\\ 
        \hline
        \multirow{6}{*}{\rotatebox[origin=c]{90}{ Static }} & \customstrut{2.5ex} Ours w/o $\loss_{\text{bone}}$  &0.8942 &0.4702 &0.8560 &0.0143 &0.1378 & 0.0724\\ 
                                                                  & \customstrut{2.5ex} Ours w/o $\loss_{\text{rigid}}$        &1.2891 &0.6758 &0.7552 &0.0158 &2.2540 & 0.1145\\
                                                                  & \customstrut{2.5ex} Ours w/o $\loss_{\text{soft}}$            &0.9282 &0.5077 &0.8567 &0.0158 &0.0460 & 0.0647\\
                                                                  & \customstrut{2.5ex} Ours w/o $\loss_{\text{lip}}$                    &0.9178 &0.4804 &0.8486 &0.0144 &0.1338 & 0.0764\\
                                                                  & \customstrut{2.5ex} Ours                                  &0.8975 &0.4721 &0.8546 &0.0143 &0.1349 & 0.0785\\ 
        \hline
        \multirow{6}{*}{\rotatebox[origin=c]{90}{ Dynamic}} & \customstrut{2.5ex} Ours w/o $\loss_{\text{bone}}$    &0.7237 &0.3182 &0.9396 &0.0070 &0.0883 & 0.0557\\ 
                                                                   & \customstrut{2.5ex} Ours w/o $\loss_{\text{rigid}}$        &1.0632 &0.4918 &0.8407 &0.0077 &2.1084 & 0.0866\\
                                                                   & \customstrut{2.5ex} Ours w/o $\loss_{\text{soft}}$           &0.7815 &0.3723 &0.9059 &0.0086 &0.0296 & 0.0400\\
                                                                   & \customstrut{2.5ex} Ours w/o $\loss_{\text{lip}}$                    &0.7751 &0.3399 &0.9270 &0.0071 &0.0915 & 0.0612\\
                                                                   & \customstrut{2.5ex} Ours                                 &0.7248 &0.3235 &0.9354 &0.0070 &0.0915 & 0.0638\\ 
        \bottomrule
    \end{tabularx}
    \label{tab:ablation}
\end{table*}

\paragraph{Ablation.}
There are two main integral loss terms in our learning framework, the rigidity loss $\loss_\text{rigid}$~(\Eqn{rigid}) 
and the soft loss $\loss_\text{soft}$~(\Eqn{soft}).
In \Tab{ablation} and \Fig{ablation}, we show the ablation studies of the
reconstructions denoted as {"Ours w/o $\loss_\text{rigid}$"} and {"Ours w/o $\loss_\text{soft}$"} to assess the importance of each loss term. 
For both the quantitative and qualitative measurements, removing
one loss term results in a severe performance decrease, which confirms
both loss terms contribute to higher reconstruction accuracy and better physical plausibility. 
Specifically, $\loss_\text{rigid}$ helps to enforce the rigid movement of the bone, giving better jaw rigidity and skull fixation metrics. This leads to better accuracy since the learned actuation mechanisms are more compatible with the rigid jaw kinematics that are strictly enforced during simulation.
$\loss_\text{soft}$ gives birth to two merits. First, it helps the model learn plausible soft tissue deformation (see Actuations in \Fig{ablation}). Second, it builds the connection between the bone and the skin, therefore being able to gradually drag the jaw to the physically plausible places during training. Therefore, {"Ours w/o $\loss_\text{soft}$"} fails to infer the jaw kinematics and always produces fixed jaw position (see Bones in \Fig{ablation}). This is why its jaw rigidity metrics and skull fixation are better than our full model (\Tab{ablation}).

\begin{figure}
    \includegraphics[width=1.0\linewidth]{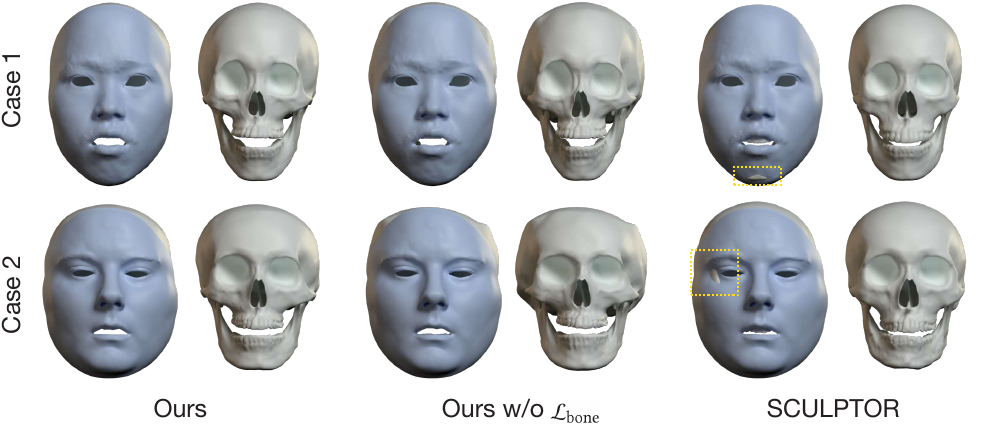}
    \caption{Qualitative comparison of different methods in terms of anatomy.}
    \label{fig:SCULPTORCOM}
\end{figure}

We further evaluate the impact of our bone shape loss $\loss_{\text{bone}}$ by training a comparative model without it, denoted as {"Ours w/o $\loss_{\text{bone}}$"}. \review{As shown in \Tab{ablation}, "Ours w/o $\loss_{\text{bone}}$" performs slightly better due to the fewer constraints on the bone shape. However, this relaxation adversely affects the realism of the bone structure (see \Tab{anatomy} and \Fig{SCULPTORCOM}), highlighting the crucial role of $\loss_{\text{bone}}$ in constraining the overall bone shape.}
Finally, our Lipschitz regularization $\loss_{\text{lip}}$ leads to better reconstruction accuracy without spoiling other metrics (see {"Ours w/o $\loss_{\text{lip}}$"} vs. "Ours" in \Tab{ablation} and \Tab{anatomy}).

\begin{figure}
    \includegraphics[width=1.0\linewidth]{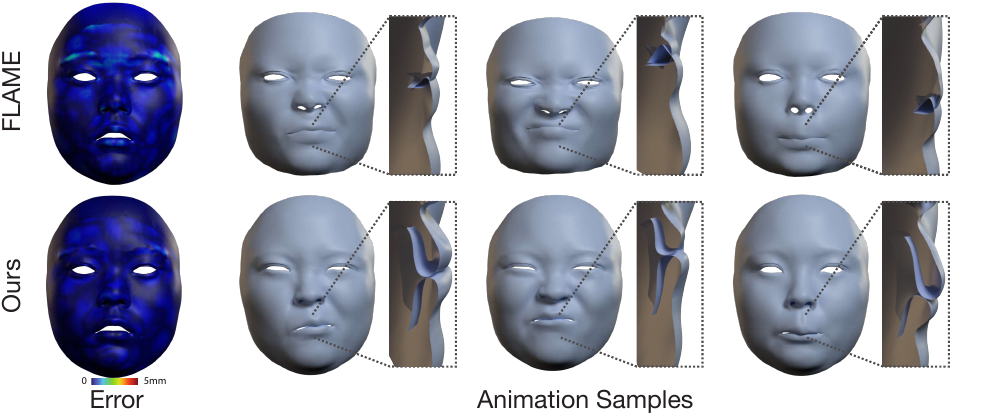}
    \caption{Qualitative comparison with FLAME. On the left, we show the fitting error maps. On the right, we show the animation samples with the mouth cut-away on the side of each frame. }
    \label{fig:FLAMECOM}
\end{figure}

\paragraph{Comparison.}
As the first generalized physical face model, we benchmark our method against the closely related FLAME model \cite{FLAME}, focusing on fitting accuracy. Recognizing that FLAME optimizes using scan-to-mesh distance, we adopt the same metric, replacing the vertex-to-vertex energy in our fitting scheme with scan-to-mesh distance, to ensure an equitable comparison. Beyond our static test set, we have compiled an additional test set of comparable size from the FACESCAPE dataset \cite{yang2020facescape}. \review{Our method is compared against two configurations of FLAME: FLAME-100, which uses the first 100 dimensions of the identity code (matching our parameterization), and FLAME-300, which employs the full space.}
Results presented in \Tab{comparison} indicate that our method attains accuracy on par with FLAME across both datasets. A significant advantage of our model, however, lies in its ability to support additional physical effects, such as collision handling—areas where traditional methods often falter. \Fig{FLAMECOM} provides a qualitative comparison to highlight this capability.

In addition, we conduct a comparison of our model with SCULPTOR, specifically focusing on the issue of bone-skin penetration. Our experiments reveal that fitting SCULPTOR to a neutral scan frequently results in bone-skin penetration problems. However, thanks to our elastic regularization ($\loss_{\text{ereg}}$), our model effectively mitigates these issues. At the same time, it maintains comparable bone shape (see \Fig{SCULPTORCOM} and \Tab{anatomy}). 

Finally, our method is still applicable in Yang et al.'s scenario when the material space is provided \cite{yang2023ExprAndStyle}. To show this, we integrate our unique loss functions—$\loss_{\text{skin}}$, $\loss_{\text{rigid}}$, and $\loss_{\text{fix}}$—into Yang et al.’s model architecture, while still adhering to their established regularization and training protocols.
\Tab{com_yang} shows the evaluation on their test set.
Our method not only achieves comparable accuracy, but it also offers significantly improved scalability. For instance, whereas Yang et al. required six GPUs and nearly 2 days to train models for six identities, our methodology can be efficiently implemented on a single GPU, enabling training for over 300 identities in just one day.

\begin{table}
    \caption{Scan-to-mesh distance of different models on different datasets. \review{"FLAME-100" is the FLAME model \cite{FLAME} using the first 100 dimensions of the identity code, whereas "FLAME-300" employs the full space. }}
    \begin{tabularx}{0.48\textwidth}{p{2.2cm}|>{\raggedleft\arraybackslash}X|>{\raggedleft\arraybackslash}X|>{\raggedleft\arraybackslash}X}
            \toprule
            \customstrut{2.5ex}{\textbf{Datasets}}&Ours*              & FLAME-100 & FLAME-300      \\ \hline
            \customstrut{2.5ex}{Our Dataset}     &0.3487             &0.4858     &0.4318           \\ 
            \customstrut{2.5ex}{FaceScape}       &0.4713             &0.6355     &0.5596           \\
            \bottomrule
        \end{tabularx}
        \label{tab:comparison}
    \end{table}

\begin{table}
    \caption{Anatomical evaluation of different models on neutral scans.}
    \begin{tabularx}{0.48\textwidth}{p{2.2cm}|>{\raggedleft\arraybackslash}X|>{\raggedleft\arraybackslash}X}
            \toprule
            \customstrut{2.5ex}{\textbf{Methods}}               &\textbf{Penetration Pairs $\downarrow$}  &\textbf{Bone Fidelity $\downarrow$}     \\ \hline
            \customstrut{2.5ex}{Ours w/o $\loss_{\text{bone}}$ }    &0                               & 1.9979         \\
            \customstrut{2.5ex}{Ours w/o $\loss_{\text{lip}}$ }                 &0                               & 1.3124         \\ 
            \customstrut{2.5ex}{Ours}                           &0                               & 1.3164         \\
            \customstrut{2.5ex}{Sculptor}                       &2628                            & 0              \\
            \bottomrule
        \end{tabularx}
        \label{tab:anatomy}
\end{table}

\begin{table}
    \caption{Quantitative Comparison with \cite{yang2023ExprAndStyle}.}
    \begin{tabularx}{0.48\textwidth}{p{2.2cm}|>{\raggedleft\arraybackslash}X|>{\raggedleft\arraybackslash}X}
            \toprule
             \textbf{Metric}           &\textbf{Yang et al}        &\textbf{Ours}     \\ \hline
            \customstrut{2.5ex}{V2V}    &0.3849                                  & 0.3813         \\
            \bottomrule
        \end{tabularx}
        \label{tab:com_yang}
\end{table}

\section{Conclusion}
\label{sec:conclusion}

We present a new model for physics-based facial animation that is trained on real data of hundreds of identities performing various expressions, and as such is extremely generalizable and can be fit to new unseen identities at runtime.  As a result, we propose a very convenient way to generate actor-specific physical face animation without any manual model setup.  This is possible due to our two main contributions: an approach for simulation-free learning where neural networks are trained to produce deformations that are compatible with physical simulation but without requiring simulation in the training loop, and a material space morphing method that can predict actor-specific skin, bones and soft-tissue volumes automatically.  These contributions are the key to being able to train on such a large dataset efficiently, providing the generalizability needed for fitting to new identities.

\begin{figure}
    \centering
    \includegraphics[width=1.0\linewidth]{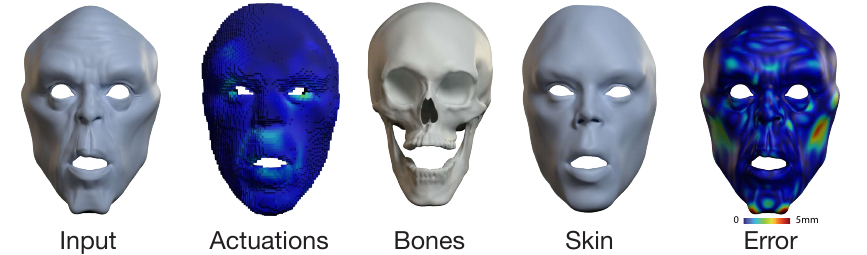}
    \caption{\review{Failure case of our method. When fitting to a non-human character, the model struggles to accurately reconstruct the geometry. We use the scan-to-mesh distance as the metric to fit and evaluate the model.}}
    \label{fig:failure_case}
\end{figure}

In terms of limitations, we note that we do not attempt to accurately model the inside of the mouth, in part because the training dataset does not accurately track this region.  As such, we ignore the teeth region on the anatomy model.  Also, while we endeavor to create constraints that produce physically-accurate animations, there is no guarantee that the learned actuations are biologically accurate, especially for unseen identities that are far from the ones seen at training time.
\review{Last, our model is designed with a focus on generality rather than specificity. As a result, capturing actor-specific facial details may not be entirely feasible with our method (see the last row of \Fig{scan_reconstruction}). Additionally, it is worth noting that when applied to non-human characters, the model may struggle to accurately reconstruct the geometry (see \Fig{failure_case}).}

Nevertheless, we demonstrate the success of our trained model with very plausible applications of fitting to new face scans, fitting to face images, animating with physical effects, animation retargeting and identity interpolation.  We additionally provide a detailed evaluation of our method.  We hope that our work will help to democratize physics-based facial animation for many applications.

\begin{acks}

We thank the anonymous reviewers for their constructive comments.
The work is supported by the 
{Swiss National Science Foundation} under Grant No.: {200021\_197136}.
\end{acks}

\bibliographystyle{ACM-Reference-Format}
\bibliography{reference}


\begin{thebibliography}{42}


\ifx \showCODEN    \undefined \def \showCODEN     #1{\unskip}     \fi
\ifx \showDOI      \undefined \def \showDOI       #1{#1}\fi
\ifx \showISBNx    \undefined \def \showISBNx     #1{\unskip}     \fi
\ifx \showISBNxiii \undefined \def \showISBNxiii  #1{\unskip}     \fi
\ifx \showISSN     \undefined \def \showISSN      #1{\unskip}     \fi
\ifx \showLCCN     \undefined \def \showLCCN      #1{\unskip}     \fi
\ifx \shownote     \undefined \def \shownote      #1{#1}          \fi
\ifx \showarticletitle \undefined \def \showarticletitle #1{#1}   \fi
\ifx \showURL      \undefined \def \showURL       {\relax}        \fi
\providecommand\bibfield[2]{#2}
\providecommand\bibinfo[2]{#2}
\providecommand\natexlab[1]{#1}
\providecommand\showeprint[2][]{arXiv:#2}

\bibitem[Bao et~al\mbox{.}(2018)]%
        {Bao19}
\bibfield{author}{\bibinfo{person}{Michael Bao}, \bibinfo{person}{Matthew Cong}, \bibinfo{person}{Stéphane Grabli}, {and} \bibinfo{person}{Ronald Fedkiw}.} \bibinfo{year}{2018}\natexlab{}.
\newblock \showarticletitle{{High-Quality Face Capture Using Anatomical Muscles}}.
\newblock \bibinfo{journal}{\emph{Proceedings of the IEEE/CVF Conference on Computer Vision and Pattern Recognition (CVPR)}} (\bibinfo{date}{12} \bibinfo{year}{2018}).
\newblock
\urldef\tempurl%
\url{http://arxiv.org/abs/1812.02836}
\showURL{%
\tempurl}


\bibitem[Blanz and Vetter(1999)]%
        {Blanz1999}
\bibfield{author}{\bibinfo{person}{Volker Blanz} {and} \bibinfo{person}{Thomas Vetter}.} \bibinfo{year}{1999}\natexlab{}.
\newblock \showarticletitle{A morphable model for the synthesis of 3D faces}. In \bibinfo{booktitle}{\emph{Siggraph}}, Vol.~\bibinfo{volume}{99}. \bibinfo{pages}{187--194}.
\newblock


\bibitem[Bouaziz et~al\mbox{.}(2014)]%
        {Bouaziz2014ProjectiveSimulation}
\bibfield{author}{\bibinfo{person}{Sofien Bouaziz}, \bibinfo{person}{Sebastian Martin}, \bibinfo{person}{Tiantian Liu}, \bibinfo{person}{Ladislav Kavan}, {and} \bibinfo{person}{Mark Pauly}.} \bibinfo{year}{2014}\natexlab{}.
\newblock \showarticletitle{{Projective dynamics: fusing constraint projections for fast simulation}}.
\newblock \bibinfo{journal}{\emph{ACM Transactions on Graphics}} \bibinfo{volume}{33}, \bibinfo{number}{4} (\bibinfo{date}{7} \bibinfo{year}{2014}), \bibinfo{pages}{1--11}.
\newblock
\showISSN{0730-0301}
\urldef\tempurl%
\url{https://doi.org/10.1145/2601097.2601116}
\showDOI{\tempurl}


\bibitem[Chai et~al\mbox{.}(2022)]%
        {REALY}
\bibfield{author}{\bibinfo{person}{Zenghao Chai}, \bibinfo{person}{Haoxian Zhang}, \bibinfo{person}{Jing Ren}, \bibinfo{person}{Di Kang}, \bibinfo{person}{Zhengzhuo Xu}, \bibinfo{person}{Xuefei Zhe}, \bibinfo{person}{Chun Yuan}, {and} \bibinfo{person}{Linchao Bao}.} \bibinfo{year}{2022}\natexlab{}.
\newblock \showarticletitle{REALY: Rethinking the Evaluation of 3D Face Reconstruction}. In \bibinfo{booktitle}{\emph{Proceedings of the European Conference on Computer Vision (ECCV)}}.
\newblock


\bibitem[Chandran et~al\mbox{.}(2020)]%
        {chandran2020semanticface}
\bibfield{author}{\bibinfo{person}{Prashanth Chandran}, \bibinfo{person}{Derek Bradley}, \bibinfo{person}{Markus Gross}, {and} \bibinfo{person}{Thabo Beeler}.} \bibinfo{year}{2020}\natexlab{}.
\newblock \showarticletitle{{Semantic Deep Face Models}}. In \bibinfo{booktitle}{\emph{2020 International Conference on 3D Vision (3DV)}}. \bibinfo{publisher}{IEEE}, \bibinfo{pages}{345--354}.
\newblock
\showISBNx{978-1-7281-8128-8}
\urldef\tempurl%
\url{https://doi.org/10.1109/3DV50981.2020.00044}
\showDOI{\tempurl}


\bibitem[Chandran et~al\mbox{.}(2022)]%
        {Chandran2022Karacast}
\bibfield{author}{\bibinfo{person}{Prashanth Chandran}, \bibinfo{person}{Lo\"{\i}c Ciccone}, \bibinfo{person}{Markus Gross}, {and} \bibinfo{person}{Derek Bradley}.} \bibinfo{year}{2022}\natexlab{}.
\newblock \showarticletitle{Local Anatomically-Constrained Facial Performance Retargeting}.
\newblock \bibinfo{journal}{\emph{ACM Trans. Graph.}} \bibinfo{volume}{41}, \bibinfo{number}{4}, Article \bibinfo{articleno}{168} (\bibinfo{date}{jul} \bibinfo{year}{2022}).
\newblock


\bibitem[Chandran and Zoss(2023)]%
        {chandran2023anatomically}
\bibfield{author}{\bibinfo{person}{Prashanth Chandran} {and} \bibinfo{person}{Gaspard Zoss}.} \bibinfo{year}{2023}\natexlab{}.
\newblock \bibinfo{title}{Anatomically Constrained Implicit Face Models}.
\newblock
\newblock
\showeprint[arxiv]{2312.07538}~[cs.GR]


\bibitem[Chandran et~al\mbox{.}(2023)]%
        {chandran2023continuous}
\bibfield{author}{\bibinfo{person}{Prashanth Chandran}, \bibinfo{person}{Gaspard Zoss}, \bibinfo{person}{Paulo Gotardo}, {and} \bibinfo{person}{Derek Bradley}.} \bibinfo{year}{2023}\natexlab{}.
\newblock \showarticletitle{Continuous Landmark Detection With 3D Queries}. In \bibinfo{booktitle}{\emph{Proceedings of the IEEE/CVF Conference on Computer Vision and Pattern Recognition}}. \bibinfo{pages}{16858--16867}.
\newblock


\bibitem[Choi et~al\mbox{.}(2022)]%
        {Choi2022}
\bibfield{author}{\bibinfo{person}{Byungkuk Choi}, \bibinfo{person}{Haekwang Eom}, \bibinfo{person}{Benjamin Mouscadet}, \bibinfo{person}{Stephen Cullingford}, \bibinfo{person}{Kurt Ma}, \bibinfo{person}{Stefanie Gassel}, \bibinfo{person}{Suzi Kim}, \bibinfo{person}{Andrew Moffat}, \bibinfo{person}{Millicent Maier}, \bibinfo{person}{Marco Revelant}, \bibinfo{person}{Joe Letteri}, {and} \bibinfo{person}{Karan Singh}.} \bibinfo{year}{2022}\natexlab{}.
\newblock \showarticletitle{Animatomy: An Animator-Centric, Anatomically Inspired System for 3D Facial Modeling, Animation and Transfer}. In \bibinfo{booktitle}{\emph{SIGGRAPH Asia 2022 Conference Papers}}. \bibinfo{publisher}{Association for Computing Machinery}, Article \bibinfo{articleno}{16}, \bibinfo{numpages}{9}~pages.
\newblock
\showISBNx{9781450394703}


\bibitem[Dai et~al\mbox{.}(2019)]%
        {LYHM}
\bibfield{author}{\bibinfo{person}{Hang Dai}, \bibinfo{person}{Nick Pears}, \bibinfo{person}{William Smith}, {and} \bibinfo{person}{Christian Duncan}.} \bibinfo{year}{2019}\natexlab{}.
\newblock \showarticletitle{Statistical Modeling of Craniofacial Shape and Texture}.
\newblock \bibinfo{journal}{\emph{International Journal of Computer Vision}} (\bibinfo{year}{2019}).
\newblock


\bibitem[Danecek et~al\mbox{.}(2022)]%
        {EMOCA:CVPR:2021}
\bibfield{author}{\bibinfo{person}{Radek Danecek}, \bibinfo{person}{Michael~J. Black}, {and} \bibinfo{person}{Timo Bolkart}.} \bibinfo{year}{2022}\natexlab{}.
\newblock \showarticletitle{{EMOCA}: {E}motion Driven Monocular Face Capture and Animation}. In \bibinfo{booktitle}{\emph{Conference on Computer Vision and Pattern Recognition (CVPR)}}. \bibinfo{pages}{20311--20322}.
\newblock


\bibitem[Egger et~al\mbox{.}(2020)]%
        {Egger20}
\bibfield{author}{\bibinfo{person}{Bernhard Egger}, \bibinfo{person}{William A.~P. Smith}, \bibinfo{person}{Ayush Tewari}, \bibinfo{person}{Stefanie Wuhrer}, \bibinfo{person}{Michael Zollhoefer}, \bibinfo{person}{Thabo Beeler}, \bibinfo{person}{Florian Bernard}, \bibinfo{person}{Timo Bolkart}, \bibinfo{person}{Adam Kortylewski}, \bibinfo{person}{Sami Romdhani}, \bibinfo{person}{Christian Theobalt}, \bibinfo{person}{Volker Blanz}, {and} \bibinfo{person}{Thomas Vetter}.} \bibinfo{year}{2020}\natexlab{}.
\newblock \showarticletitle{{3D Morphable Face Models—Past, Present, and Future}}.
\newblock \bibinfo{journal}{\emph{ACM Transactions on Graphics}} \bibinfo{volume}{39}, \bibinfo{number}{5} (\bibinfo{date}{10} \bibinfo{year}{2020}), \bibinfo{pages}{1--38}.
\newblock
\showISSN{0730-0301}
\urldef\tempurl%
\url{https://doi.org/10.1145/3395208}
\showDOI{\tempurl}


\bibitem[Feng et~al\mbox{.}(2021)]%
        {Feng:SIGGRAPH:2021}
\bibfield{author}{\bibinfo{person}{Yao Feng}, \bibinfo{person}{Haiwen Feng}, \bibinfo{person}{Michael~J. Black}, {and} \bibinfo{person}{Timo Bolkart}.} \bibinfo{year}{2021}\natexlab{}.
\newblock \showarticletitle{Learning an Animatable Detailed {3D} Face Model from In-the-Wild Images}.
\newblock \bibinfo{journal}{\emph{ACM Transactions on Graphics (ToG), Proc. SIGGRAPH}} \bibinfo{volume}{40}, \bibinfo{number}{4} (\bibinfo{date}{Aug.} \bibinfo{year}{2021}), \bibinfo{pages}{88:1--88:13}.
\newblock


\bibitem[Filntisis et~al\mbox{.}(2022)]%
        {filntisis2022visual}
\bibfield{author}{\bibinfo{person}{Panagiotis~P. Filntisis}, \bibinfo{person}{George Retsinas}, \bibinfo{person}{Foivos Paraperas-Papantoniou}, \bibinfo{person}{Athanasios Katsamanis}, \bibinfo{person}{Anastasios Roussos}, {and} \bibinfo{person}{Petros Maragos}.} \bibinfo{year}{2022}\natexlab{}.
\newblock \showarticletitle{Visual Speech-Aware Perceptual 3D Facial Expression Reconstruction from Videos}.
\newblock \bibinfo{journal}{\emph{arXiv preprint arXiv:2207.11094}} (\bibinfo{year}{2022}).
\newblock


\bibitem[Gecer et~al\mbox{.}(2019)]%
        {Gecer_2019_CVPR}
\bibfield{author}{\bibinfo{person}{Baris Gecer}, \bibinfo{person}{Stylianos Ploumpis}, \bibinfo{person}{Irene Kotsia}, {and} \bibinfo{person}{Stefanos Zafeiriou}.} \bibinfo{year}{2019}\natexlab{}.
\newblock \showarticletitle{GANFIT: Generative Adversarial Network Fitting for High Fidelity 3D Face Reconstruction}. In \bibinfo{booktitle}{\emph{The IEEE Conference on Computer Vision and Pattern Recognition (CVPR)}}.
\newblock


\bibitem[Gecer et~al\mbox{.}(2021)]%
        {gecer2021fast}
\bibfield{author}{\bibinfo{person}{Baris Gecer}, \bibinfo{person}{Stylianos Ploumpis}, \bibinfo{person}{Irene Kotsia}, {and} \bibinfo{person}{Stefanos~P Zafeiriou}.} \bibinfo{year}{2021}\natexlab{}.
\newblock \showarticletitle{Fast-GANFIT: Generative Adversarial Network for High Fidelity 3D Face Reconstruction}.
\newblock \bibinfo{journal}{\emph{IEEE Transactions on Pattern Analysis and Machine Intelligence}} (\bibinfo{year}{2021}).
\newblock


\bibitem[Gruber et~al\mbox{.}(2020)]%
        {Gruber2020}
\bibfield{author}{\bibinfo{person}{Aurel Gruber}, \bibinfo{person}{Marco Fratarcangeli}, \bibinfo{person}{Gaspard Zoss}, \bibinfo{person}{Roman Cattaneo}, \bibinfo{person}{Thabo Beeler}, \bibinfo{person}{Markus Gross}, {and} \bibinfo{person}{Derek Bradley}.} \bibinfo{year}{2020}\natexlab{}.
\newblock \showarticletitle{{Interactive Sculpting of Digital Faces Using an Anatomical Modeling Paradigm}}.
\newblock \bibinfo{journal}{\emph{Computer Graphics Forum}} (\bibinfo{year}{2020}), \bibinfo{pages}{93--102}.
\newblock
\showISSN{1467-8659}
\urldef\tempurl%
\url{https://doi.org/10.1111/cgf.14071}
\showDOI{\tempurl}


\bibitem[Hendrycks and Gimpel(2016)]%
        {hendrycks2016gaussian}
\bibfield{author}{\bibinfo{person}{Dan Hendrycks} {and} \bibinfo{person}{Kevin Gimpel}.} \bibinfo{year}{2016}\natexlab{}.
\newblock \showarticletitle{Gaussian error linear units (gelus)}.
\newblock \bibinfo{journal}{\emph{arXiv preprint arXiv:1606.08415}} (\bibinfo{year}{2016}).
\newblock


\bibitem[Ichim et~al\mbox{.}(2017)]%
        {ichim2017phace}
\bibfield{author}{\bibinfo{person}{Alexandru-Eugen Ichim}, \bibinfo{person}{Petr Kadle{\v{c}}ek}, \bibinfo{person}{Ladislav Kavan}, {and} \bibinfo{person}{Mark Pauly}.} \bibinfo{year}{2017}\natexlab{}.
\newblock \showarticletitle{{Phace: Physics-based Face Modeling and Animation}}.
\newblock \bibinfo{journal}{\emph{ACM Transactions on Graphics}} \bibinfo{volume}{36}, \bibinfo{number}{4} (\bibinfo{date}{7} \bibinfo{year}{2017}), \bibinfo{pages}{1--14}.
\newblock
\showISSN{0730-0301}
\urldef\tempurl%
\url{https://doi.org/10.1145/3072959.3073664}
\showDOI{\tempurl}


\bibitem[Kingma and Ba(2015)]%
        {kingma2015adam}
\bibfield{author}{\bibinfo{person}{Diederik~P. Kingma} {and} \bibinfo{person}{Jimmy Ba}.} \bibinfo{year}{2015}\natexlab{}.
\newblock \showarticletitle{{Adam: A Method for Stochastic Optimization}}.
\newblock \bibinfo{journal}{\emph{3rd International Conference on Learning Representations, ICLR 2015, San Diego, CA, USA, May 7-9, 2015, Conference Track Proceedings}} (\bibinfo{date}{12} \bibinfo{year}{2015}).
\newblock
\urldef\tempurl%
\url{http://arxiv.org/abs/1412.6980}
\showURL{%
\tempurl}


\bibitem[Kirschstein et~al\mbox{.}(2023)]%
        {kirschstein2023nersemble}
\bibfield{author}{\bibinfo{person}{Tobias Kirschstein}, \bibinfo{person}{Shenhan Qian}, \bibinfo{person}{Simon Giebenhain}, \bibinfo{person}{Tim Walter}, {and} \bibinfo{person}{Matthias Nie{\ss}ner}.} \bibinfo{year}{2023}\natexlab{}.
\newblock \showarticletitle{Nersemble: Multi-view radiance field reconstruction of human heads}.
\newblock \bibinfo{journal}{\emph{ACM Transactions on Graphics (TOG)}} \bibinfo{volume}{42}, \bibinfo{number}{4} (\bibinfo{year}{2023}), \bibinfo{pages}{1--14}.
\newblock


\bibitem[Kl\'{a}r et~al\mbox{.}(2020)]%
        {klar2020Shapetargeting}
\bibfield{author}{\bibinfo{person}{Gergely Kl\'{a}r}, \bibinfo{person}{Andrew Moffat}, \bibinfo{person}{Ken Museth}, {and} \bibinfo{person}{Eftychios Sifakis}.} \bibinfo{year}{2020}\natexlab{}.
\newblock \showarticletitle{Shape Targeting: A Versatile Active Elasticity Constitutive Model}. In \bibinfo{booktitle}{\emph{ACM SIGGRAPH 2020 Talks}} (Virtual Event, USA) \emph{(\bibinfo{series}{SIGGRAPH '20})}. \bibinfo{publisher}{Association for Computing Machinery}, \bibinfo{address}{New York, NY, USA}, Article \bibinfo{articleno}{59}, \bibinfo{numpages}{2}~pages.
\newblock
\showISBNx{9781450379717}
\urldef\tempurl%
\url{https://doi.org/10.1145/3388767.3407379}
\showDOI{\tempurl}


\bibitem[Li et~al\mbox{.}(2023)]%
        {li2023efficient}
\bibfield{author}{\bibinfo{person}{Bo Li}, \bibinfo{person}{Lingchen Yang}, {and} \bibinfo{person}{Barbara Solenthaler}.} \bibinfo{year}{2023}\natexlab{}.
\newblock \showarticletitle{Efficient Incremental Potential Contact for Actuated Face Simulation}.
\newblock In \bibinfo{booktitle}{\emph{SIGGRAPH Asia 2023 Technical Communications}}. \bibinfo{pages}{1--4}.
\newblock


\bibitem[Li et~al\mbox{.}(2017)]%
        {FLAME}
\bibfield{author}{\bibinfo{person}{Tianye Li}, \bibinfo{person}{Timo Bolkart}, \bibinfo{person}{Michael.~J. Black}, \bibinfo{person}{Hao Li}, {and} \bibinfo{person}{Javier Romero}.} \bibinfo{year}{2017}\natexlab{}.
\newblock \showarticletitle{Learning a model of facial shape and expression from {4D} scans}.
\newblock \bibinfo{journal}{\emph{ACM Transactions on Graphics, (Proc. SIGGRAPH Asia)}} \bibinfo{volume}{36}, \bibinfo{number}{6} (\bibinfo{year}{2017}).
\newblock


\bibitem[Liu et~al\mbox{.}(2022)]%
        {liu2022learning}
\bibfield{author}{\bibinfo{person}{Hsueh-Ti~Derek Liu}, \bibinfo{person}{Francis Williams}, \bibinfo{person}{Alec Jacobson}, \bibinfo{person}{Sanja Fidler}, {and} \bibinfo{person}{Or Litany}.} \bibinfo{year}{2022}\natexlab{}.
\newblock \showarticletitle{Learning Smooth Neural Functions via Lipschitz Regularization}.
\newblock \bibinfo{journal}{\emph{arXiv preprint arXiv:2202.08345}} (\bibinfo{year}{2022}).
\newblock


\bibitem[Morales et~al\mbox{.}(2020)]%
        {Morales2020SurveyO3}
\bibfield{author}{\bibinfo{person}{Araceli Morales}, \bibinfo{person}{Gemma Piella}, {and} \bibinfo{person}{Federico~M. Sukno}.} \bibinfo{year}{2020}\natexlab{}.
\newblock \showarticletitle{Survey on 3D face reconstruction from uncalibrated images}.
\newblock \bibinfo{journal}{\emph{Comput. Sci. Rev.}}  \bibinfo{volume}{40} (\bibinfo{year}{2020}), \bibinfo{pages}{100400}.
\newblock


\bibitem[Otto et~al\mbox{.}(2023)]%
        {otto2023perceptual}
\bibfield{author}{\bibinfo{person}{Christopher Otto}, \bibinfo{person}{Prashanth Chandran}, \bibinfo{person}{Gaspard Zoss}, \bibinfo{person}{Markus Gross}, \bibinfo{person}{Paulo Gotardo}, {and} \bibinfo{person}{Derek Bradley}.} \bibinfo{year}{2023}\natexlab{}.
\newblock \bibinfo{title}{A Perceptual Shape Loss for Monocular 3D Face Reconstruction}.
\newblock
\newblock
\showeprint[arxiv]{2310.19580}~[cs.CV]


\bibitem[Qiu et~al\mbox{.}(2022)]%
        {qiu2022sculptor}
\bibfield{author}{\bibinfo{person}{Zesong Qiu}, \bibinfo{person}{Yuwei Li}, \bibinfo{person}{Dongming He}, \bibinfo{person}{Qixuan Zhang}, \bibinfo{person}{Longwen Zhang}, \bibinfo{person}{Yinghao Zhang}, \bibinfo{person}{Jingya Wang}, \bibinfo{person}{Lan Xu}, \bibinfo{person}{Xudong Wang}, \bibinfo{person}{Yuyao Zhang}, {et~al\mbox{.}}} \bibinfo{year}{2022}\natexlab{}.
\newblock \showarticletitle{SCULPTOR: Skeleton-consistent face creation using a learned parametric generator}.
\newblock \bibinfo{journal}{\emph{ACM Transactions on Graphics (TOG)}} \bibinfo{volume}{41}, \bibinfo{number}{6} (\bibinfo{year}{2022}), \bibinfo{pages}{1--17}.
\newblock


\bibitem[Raissi et~al\mbox{.}(2019)]%
        {raissi2019physics}
\bibfield{author}{\bibinfo{person}{Maziar Raissi}, \bibinfo{person}{Paris Perdikaris}, {and} \bibinfo{person}{George~E Karniadakis}.} \bibinfo{year}{2019}\natexlab{}.
\newblock \showarticletitle{Physics-informed neural networks: A deep learning framework for solving forward and inverse problems involving nonlinear partial differential equations}.
\newblock \bibinfo{journal}{\emph{Journal of Computational physics}}  \bibinfo{volume}{378} (\bibinfo{year}{2019}), \bibinfo{pages}{686--707}.
\newblock


\bibitem[Sifakis et~al\mbox{.}(2005)]%
        {Sifakis05}
\bibfield{author}{\bibinfo{person}{Eftychios Sifakis}, \bibinfo{person}{Igor Neverov}, {and} \bibinfo{person}{Ronald Fedkiw}.} \bibinfo{year}{2005}\natexlab{}.
\newblock \showarticletitle{{Automatic determination of facial muscle activations from sparse motion capture marker data}}.
\newblock \bibinfo{journal}{\emph{ACM Transactions on Graphics}} \bibinfo{volume}{24}, \bibinfo{number}{3} (\bibinfo{date}{7} \bibinfo{year}{2005}), \bibinfo{pages}{417--425}.
\newblock
\showISSN{0730-0301}
\urldef\tempurl%
\url{https://doi.org/10.1145/1073204.1073208}
\showDOI{\tempurl}


\bibitem[Sitzmann et~al\mbox{.}(2020)]%
        {siren2020}
\bibfield{author}{\bibinfo{person}{Vincent Sitzmann}, \bibinfo{person}{Julien N~P Martel}, \bibinfo{person}{Alexander~W Bergman}, \bibinfo{person}{David~B Lindell}, {and} \bibinfo{person}{Gordon Wetzstein}.} \bibinfo{year}{2020}\natexlab{}.
\newblock \showarticletitle{{Implicit Neural Representations with Periodic Activation Functions}}. In \bibinfo{booktitle}{\emph{NeurIPS 2020}}, \bibfield{editor}{\bibinfo{person}{Hugo Larochelle}, \bibinfo{person}{Marc'Aurelio Ranzato}, \bibinfo{person}{Raia Hadsell}, \bibinfo{person}{Maria-Florina Balcan}, {and} \bibinfo{person}{Hsuan-Tien Lin}} (Eds.).
\newblock
\urldef\tempurl%
\url{https://proceedings.neurips.cc/paper/2020/hash/53c04118df112c13a8c34b38343b9c10-Abstract.html}
\showURL{%
\tempurl}


\bibitem[Srinivasan et~al\mbox{.}(2021)]%
        {srinivasan2021learning}
\bibfield{author}{\bibinfo{person}{Sangeetha~Grama Srinivasan}, \bibinfo{person}{Qisi Wang}, \bibinfo{person}{Junior Rojas}, \bibinfo{person}{Gergely Kl{\'{a}}r}, \bibinfo{person}{Ladislav Kavan}, {and} \bibinfo{person}{Eftychios Sifakis}.} \bibinfo{year}{2021}\natexlab{}.
\newblock \showarticletitle{{Learning active quasistatic physics-based models from data}}.
\newblock \bibinfo{journal}{\emph{ACM Transactions on Graphics}} \bibinfo{volume}{40}, \bibinfo{number}{4} (\bibinfo{date}{8} \bibinfo{year}{2021}), \bibinfo{pages}{1--14}.
\newblock
\showISSN{0730-0301}
\urldef\tempurl%
\url{https://doi.org/10.1145/3450626.3459883}
\showDOI{\tempurl}


\bibitem[Wagner et~al\mbox{.}(2023)]%
        {SoftDECA2023}
\bibfield{author}{\bibinfo{person}{Nicolas Wagner}, \bibinfo{person}{Mario Botsch}, {and} \bibinfo{person}{Ulrich Schwanecke}.} \bibinfo{year}{2023}\natexlab{}.
\newblock \showarticletitle{SoftDECA: Computationally Efficient Physics-Based Facial Animations}. In \bibinfo{booktitle}{\emph{Proceedings of the 16th ACM SIGGRAPH Conference on Motion, Interaction and Games}} \emph{(\bibinfo{series}{MIG '23})}. \bibinfo{publisher}{Association for Computing Machinery}, \bibinfo{address}{New York, NY, USA}, Article \bibinfo{articleno}{11}, \bibinfo{numpages}{11}~pages.
\newblock
\showISBNx{9798400703935}
\urldef\tempurl%
\url{https://doi.org/10.1145/3623264.3624439}
\showDOI{\tempurl}


\bibitem[Wu et~al\mbox{.}(2016)]%
        {Wu2016}
\bibfield{author}{\bibinfo{person}{Chenglei Wu}, \bibinfo{person}{Derek Bradley}, \bibinfo{person}{Markus Gross}, {and} \bibinfo{person}{Thabo Beeler}.} \bibinfo{year}{2016}\natexlab{}.
\newblock \showarticletitle{{An anatomically-constrained local deformation model for monocular face capture}}.
\newblock \bibinfo{journal}{\emph{ACM Transactions on Graphics}} \bibinfo{volume}{35}, \bibinfo{number}{4} (\bibinfo{date}{7} \bibinfo{year}{2016}), \bibinfo{pages}{1--12}.
\newblock
\showISSN{0730-0301}
\urldef\tempurl%
\url{https://doi.org/10.1145/2897824.2925882}
\showDOI{\tempurl}


\bibitem[Xu and Yang(2015)]%
        {xu2015human}
\bibfield{author}{\bibinfo{person}{Ming Xu} {and} \bibinfo{person}{James Yang}.} \bibinfo{year}{2015}\natexlab{}.
\newblock \showarticletitle{Human facial soft tissue thickness and mechanical properties: a literature review}. In \bibinfo{booktitle}{\emph{International Design Engineering Technical Conferences and Computers and Information in Engineering Conference}}, Vol.~\bibinfo{volume}{57045}. American Society of Mechanical Engineers, \bibinfo{pages}{V01AT02A045}.
\newblock


\bibitem[Yang et~al\mbox{.}(2020a)]%
        {FaceScape}
\bibfield{author}{\bibinfo{person}{Haotian Yang}, \bibinfo{person}{Hao Zhu}, \bibinfo{person}{Yanru Wang}, \bibinfo{person}{Mingkai Huang}, \bibinfo{person}{Qiu Shen}, \bibinfo{person}{Ruigang Yang}, {and} \bibinfo{person}{Xun Cao}.} \bibinfo{year}{2020}\natexlab{a}.
\newblock \showarticletitle{FaceScape: a Large-scale High Quality 3D Face Dataset and Detailed Riggable 3D Face Prediction}. In \bibinfo{booktitle}{\emph{Proceedings of the IEEE Conference on Computer Vision and Pattern Recognition (CVPR)}}.
\newblock


\bibitem[Yang et~al\mbox{.}(2020b)]%
        {yang2020facescape}
\bibfield{author}{\bibinfo{person}{Haotian Yang}, \bibinfo{person}{Hao Zhu}, \bibinfo{person}{Yanru Wang}, \bibinfo{person}{Mingkai Huang}, \bibinfo{person}{Qiu Shen}, \bibinfo{person}{Ruigang Yang}, {and} \bibinfo{person}{Xun Cao}.} \bibinfo{year}{2020}\natexlab{b}.
\newblock \showarticletitle{Facescape: a large-scale high quality 3d face dataset and detailed riggable 3d face prediction}. In \bibinfo{booktitle}{\emph{Proceedings of the ieee/cvf conference on computer vision and pattern recognition}}. \bibinfo{pages}{601--610}.
\newblock


\bibitem[Yang et~al\mbox{.}(2022)]%
        {yang2022implicit}
\bibfield{author}{\bibinfo{person}{Lingchen Yang}, \bibinfo{person}{Byungsoo Kim}, \bibinfo{person}{Gaspard Zoss}, \bibinfo{person}{Baran G{\"o}zc{\"u}}, \bibinfo{person}{Markus Gross}, {and} \bibinfo{person}{Barbara Solenthaler}.} \bibinfo{year}{2022}\natexlab{}.
\newblock \showarticletitle{Implicit neural representation for physics-driven actuated soft bodies}.
\newblock \bibinfo{journal}{\emph{ACM Transactions on Graphics (TOG)}} \bibinfo{volume}{41}, \bibinfo{number}{4} (\bibinfo{year}{2022}), \bibinfo{pages}{1--10}.
\newblock


\bibitem[Yang et~al\mbox{.}(2023)]%
        {yang2023ExprAndStyle}
\bibfield{author}{\bibinfo{person}{Lingchen Yang}, \bibinfo{person}{Gaspard Zoss}, \bibinfo{person}{Prashanth Chandran}, \bibinfo{person}{Paulo Gotardo}, \bibinfo{person}{Markus Gross}, \bibinfo{person}{Barbara Solenthaler}, \bibinfo{person}{Eftychios Sifakis}, {and} \bibinfo{person}{Derek Bradley}.} \bibinfo{year}{2023}\natexlab{}.
\newblock \showarticletitle{An Implicit Physical Face Model Driven by Expression and Style}. In \bibinfo{booktitle}{\emph{SIGGRAPH Asia 2023 Conference Papers}}. Article \bibinfo{articleno}{106}, \bibinfo{numpages}{12}~pages.
\newblock


\bibitem[Zhang et~al\mbox{.}(2023)]%
        {Zhang_2023_ICCV}
\bibfield{author}{\bibinfo{person}{Tianke Zhang}, \bibinfo{person}{Xuangeng Chu}, \bibinfo{person}{Yunfei Liu}, \bibinfo{person}{Lijian Lin}, \bibinfo{person}{Zhendong Yang}, \bibinfo{person}{Zhengzhuo Xu}, \bibinfo{person}{Chengkun Cao}, \bibinfo{person}{Fei Yu}, \bibinfo{person}{Changyin Zhou}, \bibinfo{person}{Chun Yuan}, {and} \bibinfo{person}{Yu Li}.} \bibinfo{year}{2023}\natexlab{}.
\newblock \showarticletitle{Accurate 3D Face Reconstruction with Facial Component Tokens}. In \bibinfo{booktitle}{\emph{Proceedings of the IEEE/CVF International Conference on Computer Vision (ICCV)}}. \bibinfo{pages}{9033--9042}.
\newblock


\bibitem[Zheng et~al\mbox{.}(2022)]%
        {zheng2022imface}
\bibfield{author}{\bibinfo{person}{Mingwu Zheng}, \bibinfo{person}{Hongyu Yang}, \bibinfo{person}{Di Huang}, {and} \bibinfo{person}{Liming Chen}.} \bibinfo{year}{2022}\natexlab{}.
\newblock \showarticletitle{Imface: A nonlinear 3d morphable face model with implicit neural representations}. In \bibinfo{booktitle}{\emph{Proceedings of the IEEE/CVF Conference on Computer Vision and Pattern Recognition}}. \bibinfo{pages}{20343--20352}.
\newblock


\bibitem[Zielonka et~al\mbox{.}(2022)]%
        {Zielonka2022TowardsMR}
\bibfield{author}{\bibinfo{person}{Wojciech Zielonka}, \bibinfo{person}{Timo Bolkart}, {and} \bibinfo{person}{Justus Thies}.} \bibinfo{year}{2022}\natexlab{}.
\newblock \showarticletitle{Towards Metrical Reconstruction of Human Faces}. In \bibinfo{booktitle}{\emph{European Conference on Computer Vision}}.
\newblock


\end{thebibliography}

\end{document}